\documentclass[acmsmall]{acmart}

\AtBeginDocument{%
	\providecommand\BibTeX{{%
			\normalfont B\kern-0.5em{\scshape i\kern-0.25em b}\kern-0.8em\TeX}}}

\setcopyright{acmcopyright}
\acmJournal{POMACS}
\acmYear{2022} \acmVolume{6} \acmNumber{2} \acmArticle{39} \acmMonth{6} \acmPrice{15.00}\acmDOI{10.1145/3530905}

\received{February 2022}
\received[revised]{March 2022}
\received[accepted]{April 2022}

\usepackage[utf8]{inputenc}
\usepackage{graphicx}
\usepackage{caption}
\usepackage{subcaption}
\usepackage{xcolor}
\usepackage{xspace}
\graphicspath{ {./figures/} }

\usepackage{verbatim} 
\usepackage{dsfont}

\DeclareMathOperator{\Var}{Var}

\DeclareMathOperator{\Cov}{Cov}

\newcommand{\Ind}[1]{\mathds{1}_{\left[ #1 \right]}}

\newcounter{actr}
{\begin{list}{(\alph{actr})}{\usecounter{actr}}}{\end{list}}

\newcounter{ictr}
{\begin{list}{(\roman{ictr})}{\usecounter{ictr}}}{\end{list}}

\newtheorem{theorem}{Theorem}
\newtheorem{lemma}{Lemma}

\newenvironment{new-proof}[1]%
{\noindent{\em Proof of #1: } }%
{\ \noindent\qed}

\newcounter{nmdthmcnt}

\newcommand{\ev}{\mathbb{E}}
\newcommand{\E}[1]{\ev\left[{#1}\right]}

\newcommand{\Prob}{\mathbb{P}}

\newcommand{\Reals}{\mathbb{R}}

\newcommand{\cA}{{\mathcal{A}}}

\newcommand{\cB}{{\mathcal{B}}}

\newcommand{\tB}{{\tilde{B}}}

\newcommand{\cE}{{\mathcal{E}}}

\newcommand{\cF}{{\mathcal{F}}}

\newcommand{\cG}{{\mathcal{G}}}

\newcommand{\bh}{{\mathbf{h}}}

\newcommand{\bi}{{\mathbf{i}}}
\newcommand{\cI}{{\mathcal{I}}}

\newcommand{\cJ}{{\mathcal{J}}}

\newcommand{\bk}{{\mathbf{k}}}

\newcommand{\tN}{{\tilde{N}}}  

\newcommand{\tp}{{\tilde{p}}}

\newcommand{\tQ}{\tilde{Q}}

\newcommand{\cS}{{\mathcal{S}}}

\newcommand{\cT}{{\mathcal{T}}}

\newcommand{\hU}{\hat{U}}

\newcommand{\hV}{\hat{V}}

\newcommand{\cV}{{\mathcal{V}}}

\newcommand{\bx}{{\mathbf{x}}}

\newcommand{\tX}{{\tilde{X}}}

\newcommand{\e}{\epsilon}

\newcommand{\hsig}{\hat{\sigma}}  
\newcommand{\hSig}{\hat{\Sigma}}  

\newcommand{\hLam}{\hat{\Lambda}}  

\theoremstyle{acmplain}
\newtheorem{assumption}{Assumption}

\begin{document}
	
\title{Tensor Completion with Nearly Linear Samples \\Given Weak Side Information}
\author{Christina Lee Yu}
\affiliation{%
	\institution{Cornell University}
	\city{Ithaca}
	\state{New York}
	\country{USA}}
\email{cleeyu@cornell.edu}
\author{Xumei Xi}
\affiliation{%
	\institution{Cornell University}
	\city{Ithaca}
	\state{New York}
	\country{USA}}
\email{xx269@cornell.edu}

\begin{abstract}
Tensor completion exhibits an interesting computational-statistical gap in terms of the number of samples needed to perform tensor estimation. While there are only $\Theta(tn)$ degrees of freedom in a $t$-order tensor with $n^t$ entries, the best known polynomial time algorithm requires $O(n^{t/2})$ samples in order to guarantee consistent estimation. In this paper, we show that weak side information is sufficient to reduce the sample complexity to $O(n)$. The side information consists of a weight vector for each of the modes which is not orthogonal to any of the latent factors along that mode; this is significantly weaker than assuming noisy knowledge of the subspaces. We provide an algorithm that utilizes this side information to produce a consistent estimator with $O(n^{1+\kappa})$ samples for any small constant $\kappa > 0$. We also provide experiments on both synthetic and real-world datasets that validate our theoretical insights.
\end{abstract}

%

\maketitle


\section{Introduction}

A tensor is a mathematical object that can be used to represent multiway data. A dataset in which each datapoint is indexed by $t$ indices can be represented by a $t$ order tensor. A 2-order tensor is simply a matrix, with each datapoint being referenced by two indices referring to the row and column. Multiway data arises in many applications. For example, image data can be represented by a 3-order tensor, with two indices referring to the pixel location, and the third index referring to the color modes of RGB. Video data could then be represented by a 4-order tensor with the 4th mode representing time. E-commerce data is also multiway, with each datapoint of interaction on the platform being associated to a user id, product id, and timestamp. As network data is naturally represented in a matrix form, data collected from monitoring a network changing over time can be represented in a 3-order tensor. Neuroimaging data involves 3D-scans that can be represented in a 3-order tensor. Microbiome studies or protein interaction data involves network data of co-ocurrence or interaction counts; this network can be collected across many different patients with different demographics, and it may be useful to represent the full data as a 3-order data to look for patterns amongst subpopulations rather than to just aggregate the data into a single matrix.

Often the tensor dataset can be very sparse due to the observation process. For example, e-commerce data is very sparse as any given user only interacts with a small subset of the products at sparse timepoints. In experimental studies, each datapoint may be costly to collect, and thus the sparsity could be limited by available budget. The observations themselves can also be noisy or corrupted due to the experimental process or mistakes in data entry. As a result, the task of tensor estimation, or learning the underlying structure given noisy and incomplete tensor data, is a significant building block in the data analysis pipeline. 

In the setting of sparse tensor completion, a critical question is how many datapoints does one need to observe (sampled uniformly at random) in order to estimate the underlying tensor structure? Consider a $t$- order data tensor with $n^t$ entries, i.e. each mode has dimension $n$. When we only observe a small subset of entries, it is impossible to guarantee recovery without imposing structure on the underlying tensor. The typical assumptions to impose are low rank and incoherence style conditions. Essentially the low rank conditions reduce the number of unknown model parameters to linear in $n$ even though the number of possible tensor entries is $n^t$ for a $t$-order tensor. The simple statistical lower bound on the sample complexity, or minimum number of observations for recovery, is thus $\Omega(n)$, as there are linear in $n$ degrees of freedom in the low-rank model. Tensor nuclear norm minimization requires $O(n^{3/2})$ observations for a $t$-order tensor, however the algorithm is not polynomial time computable as tensor nuclear norm is NP-hard to compute \cite{yuan2017incoherent, friedland2014nuclear}. The best existing polynomial time algorithms require $O(n^{t/2})$ observations for a $t$-order tensor. There is a large gap between what is polynomial time achievable and the statistical lower bound.

For a 3-order tensor, \cite{BarakMoitra16} conjectured that $\Omega(n^{3/2})$ samples are needed for polynomial time computation based on a reduction of tensor completion for a class of a rank 1 tensors to the random 3-XOR distinguishability problem. Conditioned on the hardness of random 3-XOR distinguishability, their result proves that any approach for tensor completion that relies on the sum of squares hierarchy or Rademacher complexity will require $\Omega(n^{3/2})$ samples. The fact that the class of hard instances in \cite{BarakMoitra16} are simply rank 1 tensors suggests that rank may not be a sufficient measure of complexity for tensor estimation.
As a result of the hardness conjecture, recent literature has accepted the threshold of $\Theta(n^{t/2})$ as a likely lower bound for computationally efficient algorithms, and instead has shifted attention to reducing dependence with respect to constant properties of the model, such as rank or incoherence constants.

In this paper we consider what conditions are sufficient to achieve nearly linear sample complexity by the use of auxiliary information. In the most general setting of tensor estimation, the indices of the data entries themselves are not expected to carry valuable information other than linking it to other entries associated to the same index. In particular, the distribution of the data is expected to be equivalent up to permutations of the indices. However, in reality we often have addition knowledge or side information about the indices in each mode such that the full generality of an exchangeable model is not the most appropriate model. For example, with video or imaging data, we expect the data to exhibit smoothness with respect to the pixel location and frame number, which are encoded in the indices of the associated modes. For e-commerce data, there is auxiliary data about the users and products that could relate to the latent factors. While there has been several empirical works studying the potential promise for utilizing side information in tensor completion, none of the works provide theoretical results with statistical guarantees. 
\subsection{Related Literature}

Tensor completion has been studied in the literature as a natural extension of matrix completion to higher dimensions. The approach and techniques have naturally focused around extensions of matrix completion techniques to tensor completion. The earliest approaches unfold the tensor to a matrix, and apply variations of matrix completion algorithms to the constructed matrix \cite{tomioka2010estimation, tomioka2011statistical, gandy2011tensor, liu2012tensor}.
Given a $t$-order tensor where each mode has dimension $n$, there are $2^{t} - 1$ possible unfoldings of the tensor to a matrix, each corresponding to a partition of the $t$ modes. For example, if $\tau_A$ and $\tau_B$ are disjoint non-empty subsets of $[t]$ such that $\tau_A \cup \tau_B = [t]$, the corresponding unfolding of the tensor would result in a $n^{|\tau_A|} \times n^{|\tau_B|}$ matrix, where each row would correspond to a member of the cartesian product of coordinates in the modes referenced by $\tau_A$, and each column would correspond to a member of the cartesian product of coordinates in the modes referenced by $\tau_B$. There would still be overall $n^t$ entries in the matrix, with a 1-1 mapping to the $n^t$ entries in the original tensor.
The above results utilize the fact that low rank and incoherence conditions for the original tensor result in low rank and incoherence conditions for the unfolding of the tensor to the constructed matrix.
As matrix completion algorithms are limited in their sample complexity by the maximum of the dimensions of the rows and columns, the unfoldings that minimize sample complexity are those that unfold the tensor to a $n^{\lfloor t/2 \rfloor} \times n^{\lceil t/2 \rceil}$ matrix, resulting in a sample complexity of $O(n^{\lceil t/2 \rceil})$. 


Unfolding the tensor to a matrix is limiting as the algorithm loses knowledge of the relationships amongst the rows and columns of the matrix that were exhibited in the original tensor, i.e. there are rows and columns in the unfolded matrix that share coordinates along some modes of the original tensor. 
There have subsequently been a series of works that have attempted to use tensor structure to reduce the sample complexity. The majority of results have focused on the setting of a 3rd order tensor, but many of the results can also be extended to general $t$-order tensors. 
Tensor nuclear norm minimization requires only sample complexity of $O(n^{3/2})$ for a general $t$-order tensor, but tensor nuclear norm is NP-hard to compute and thus does not lead to a polynomial time algorithm \cite{yuan2016tensor, friedland2014nuclear}. The best polynomial time algorithms require a sample complexity of $O(n^{t/2})$ for an order $t$ tensor. These results have been attained using extensions from a variety of techniques similar to the matrix completion setting, including spectral style methods \cite{MontanariSun18, xia2017polynomial}, convex relaxation via sum of squares \cite{BarakMoitra16, PotechinSteurer17}, minimizing the nonconvex objective directly via gradient descent \cite{xia2017polynomial, xia2017statistically, cai2021nonconvex} or alternating least squares \cite{jain2014provable, bhojanapalli2015new}, or iterative collaborative filtering \cite{shah2019iterative}. The naive statistical lower bound is $\Omega(n)$ as the number of unknown parameters in a low rank model grows linearly in $n$. This still leaves a large gap between the sample complexity of the best existing polynomial time algorithms and the statistical lower bound. \cite{BarakMoitra16} provides evidence for a computational-statistical gap by relating tensor completion via the Rademacher complexity and sum of squares framework to refutation of random 3-SAT. 


While the above works all consider a uniform sampling model, \cite{zhang2019} considers an active sampling scheme that achieves optimal sample efficiency of $O(n)$. Their approach requires a specific sampling scheme that aligns all the samples to guarantee that along each subspace there are entire columns of data sampled. While this result yields optimal bounds and is useful for settings where the data can be actively sampled, many applications do not allow such active control over the sampling process. \cite{cai2020provable} considers the tensor recovery problem, which allows for general measurement operators instead of only single entry observations. They prove that spectral initialization with Riemannian gradient descent can recover the underlying tensor with only $O(n r^2)$ Gaussian measurements, achieving the optimal linear dependence on $n$. Their result relies on the tensor restricted isometry property that arises from the Gaussian measurements. As such, it does not extend to entrywise observations.


The inductive matrix completion problem considers the setting when exact or partial subspace information is provided alongside the matrix completion task \cite{burkina2021inductive, ghassemi2018global, chiang2015matrix, chiang2018using, jain2013provable}. When exact subspace information is given, the degrees of freedom in the model is significantly reduced as one only needs to estimate the smaller core matrix governing the interaction between the row and column subspaces. As a result, the sample complexity reduces from a linear to logarithmic dependence on the matrix dimension $n$ \cite{jain2013provable}. \cite{chiang2015matrix} further considers the setting where noisy or partial information about the subspace is given, modeling the desired matrix as a sum of the interaction between the side information and a low rank residual matrix. However, under partial or noisy side information, the sample complexity is still linear in $n$ as the degrees of freedom in the model is still linear in $n$. There have been empirical works showing benefits of utilizing side information for tensor completion. The most common model assumes that the side information is given in the form of a subspace for each of the modes that contains the associated column subspace of each matricization associated to an unfolding of the tensor \cite{bertsimastensor, zhou2017tensor, budzinskiy2020note, nimishakavi2018inductive,chen2019collective}; this is a natural extension of the inductive matrix completion setting. Given the subspaces, since the degrees of freedom in the model no longer grows with $n$, one would expect that the sample complexity would reduce to logarithmic in $n$, although none of these papers provide formal statistical guarantees. \cite{narita2012tensor} assumes that the side information is given in the form of similarity matrices amongst indices in each mode of the tensor, which they incorporate into the algorithm via a Laplacian regularizer. \cite{lamba2016incorporating} considers a Bayesian setup in which the side information is in the form of kernel matrices that are used to construct Gaussian process priors for the latent factors. None of these above mentioned results in tensor completion with side information provide formal statistical guarantees, although the empirical results seem promising. 

We utilize a similar insight as \cite{kolda2015symmetric}, which shows that for orthogonal symmetric tensors, when all entries are observed, tensor decomposition can be computed by constructing $n \times n$ matrices with the same latent factors as the tensor. Their setting assumes that all entries are observed, and thus does not provide an algorithm or statistical guarantees for noisy and sparsely observed datasets. We extend the idea to sparsely observed tensors and beyond orthogonal symmetric tensors. Our specific algorithm uses the nearest neighbor collaborative filtering approach for matrix completion, introduced in \cite{song2016blind, li2019nearest, BorgsChayesLeeShah17, borgs2021iterative, shah2019iterative}. 



\subsection{Contributions}

Consider a $t$-order tensor with dimension $n$ along each mode, and assume we are given a sparse set of noisy observations where each entry is sampled independently with probability $p$ and observed with mean zero bounded noise. When the tensor has low orthogonal CP-rank, we assume a weak form of side information consisting of a single vector for each mode which simply lies in the column space of the associated matricization of the tensor along that mode. Furthermore suppose the side information vectors are not exactly aligned with any of the latent factors. Under these assumptions, we provide a simple polynomial time algorithm which provably outputs a consistent estimator as long as the number of observed entries is at least $\Omega(n^{1+\kappa})$ for any arbitrarily small constant $\kappa > 0$, nearly achieving the linear lower bound resulting from the degrees of freedom in the model. We extend our results beyond low orthogonal CP-rank tensors as well, providing a characterization for sufficient side information to admit nearly linear sample complexity.

To our knowledge, this is the first theoretical result for tensor completion with side information, provably showing that given weak side information the sample complexity of tensor estimation can reduce from the conjectured $n^{t/2}$ to nearly linear in $n$. The side information we assume is significantly weaker than assuming knowledge of the full subspaces, and thus is more plausible for real applciations. 
Our proposed algorithm is simple, essentially using matrix estimation techniques on constructed matrices of size $n \times n$ to learn similarities between coordinates. These similarities are used to estimate the underlying tensor via a nearest neighbor estimator. An additional benefit of our analysis is that we are able to prove that with high probability, the maximum entrywise error of our estimate decays as $\tilde{O}(\max(n^{\kappa/4},n^{-(\kappa+1)/(t+2)})$ where the expected number of observations is $n^{1+\kappa}$ for any small constant $\kappa > 0$ and the $\tilde{O}$ notation simply hides polylog factors. Our result implies an infinity norm bound on the error rather than the looser typical mean squared error bounds provided in the existing literature. We also provide experiments on both synthetic and real-world datasets that validate our theoretical insights.


\section{Preliminaries}

Consider the task of estimating a $t$-order tensor $T$ given observations $\{T^{obs}(\bi)\}_{\bi \in \Omega}$  for some $\Omega \subset [n_1] \times [n_2] \times \cdots [n_t]$ where $|\Omega|$ is significantly smaller than $n_1 n_2 \dots n_t$. Assume an additive noise model where $T^{obs}(\bi) = T(\bi) + E(\bi)$ for $\bi \in \Omega$, and $E$ is the noise matrix with independent mean zero entries. We assume a uniform Bernoulli sampling model where each entry is observed independently with probability $p$, i.e. $\Ind{\bi \in \Omega} \sim$ Bernoulli($p$).

\begin{figure} 
	\centering
	\includegraphics[width=5in]{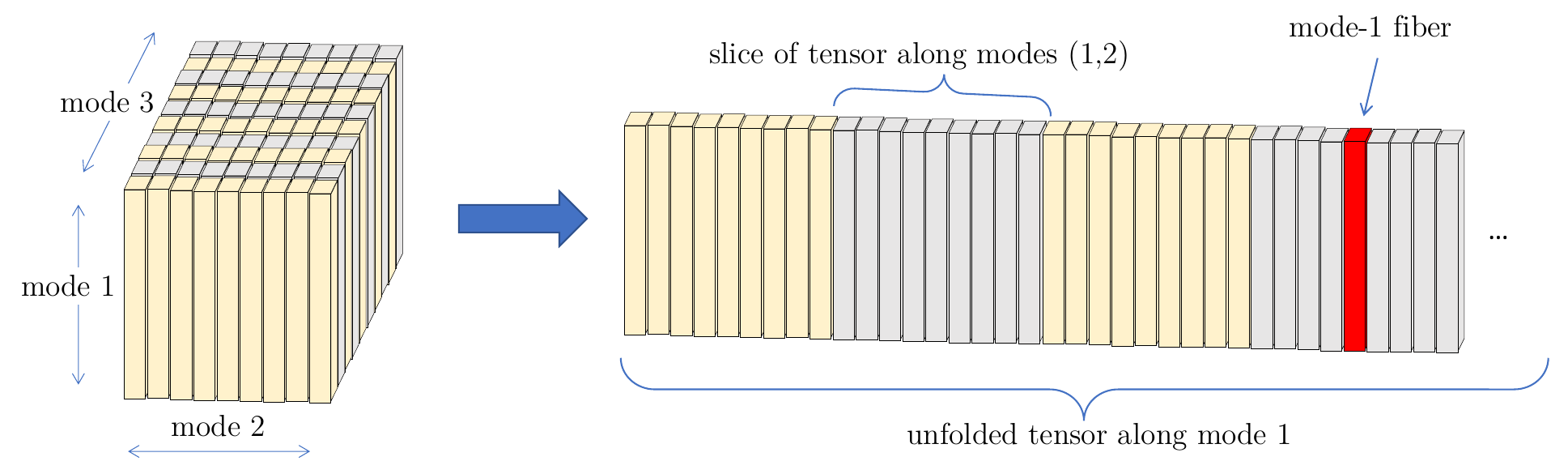}
	\caption{Depicting an unfolding of a 3rd order tensor along mode 1, denoted as $T_{(1)}$. The columns of the resulting matrix are referred to as the mode-1 fibers of the tensor. The submatrices alternating in color are referred to as slices of the tensor along modes $(1,2)$.} \label{fig:unfolding}
\end{figure}

Let $T_{(\ell)}$ denote the unfolded tensor along the $\ell$-th mode, which is a matrix of dimension $n_{\ell} \times \prod_{\ell' \in [t] \setminus \{\ell\}} n_{\ell'}$. We refer to columns of $T_{(\ell)}$ as mode-$\ell$ fibers of tensor $T$, which consists of vectors constructed by varying only the index in mode $\ell$ and fixing the indices along all other modes not equal to $\ell$. We refer to slices of the tensor along modes $(\ell,\ell')$ to be the matrix of entries resulting from varying the indices in modes $\ell$ and $\ell'$ and fixing the indices along all modes not equal to $\ell$ or $\ell'$. Figure \ref{fig:unfolding} visualizes one such unfolding of the tensor. 

\begin{figure}
	\centering
	\includegraphics[width=5in]{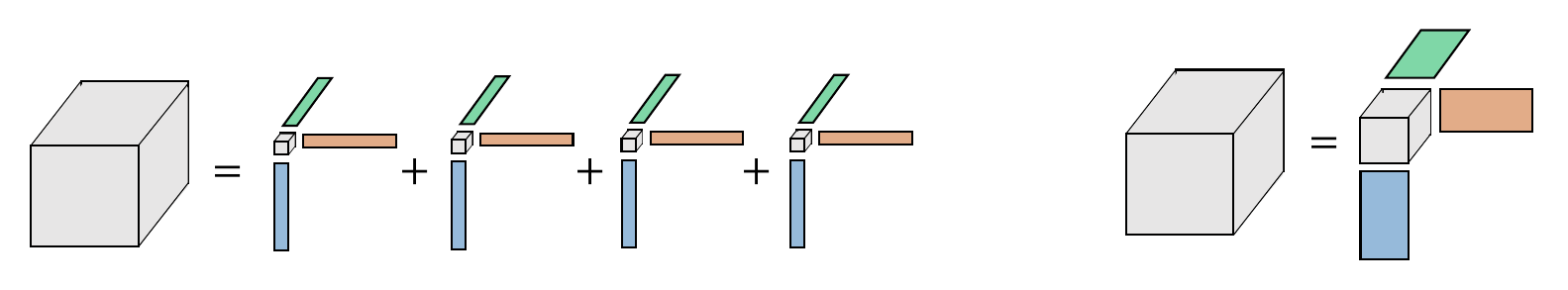}
	\caption{(Left) The tensor CP-rank admits a decomposition corresponding to the sum of $r$ rank-1 tensors. \\(Right) The Tucker rank or multilinear rank $(r_1, r_2, \dots r_t)$ admits a decomposition corresponding to a multilinear multiplication of a core tensor of dimensions $(r_1, r_2, \dots r_t)$ with latent factor matrices associated to each mode. The number of degrees of freedom in each model scales linearly with $n$ via the latent factors.}
	 \label{fig:decomposition}
\end{figure}

We must impose low dimensional structure on the underlying tensor to reasonably expect sparse tensor estimation to be feasible. Unlike in the matrix setting, there are multiple definitions of tensor rank. {\em CP-rank} is the minimum number of rank-1 tensors such that their sum is equal to the desired tensor. An {\em overcomplete} tensor is one for which the CP-rank is larger than the dimension $n$. The latent factors in the minimal rank-1 CP-decomposition may not be orthogonal.
%
The {\em Tucker rank}, or {\em multilinear rank}, is a vector $(r_1, r_2, \dots r_t)$ such that for each mode $\ell \in [t]$, $r_{\ell}$ is the dimension of the column space of $T_{(\ell)}$, the unfolded tensor along the $\ell$-th mode to a $n_{\ell} \times \prod_{i \neq \ell} n_i$ matrix. The Tucker rank is also the minimal values of $(r_1, r_2, \dots r_t)$ for which the tensor can be decomposed according to a multilinear multiplication of a core tensor $\Lambda \in \Reals^{r_1 \times r_2 \times \dots r_t}$ with latent factor matrices $Q_1 \dots Q_t$ for $Q_{\ell} \in \Reals^{n_{\ell} \times r_{\ell}}$, denoted as
\begin{align}
T = (Q_1 \otimes \cdots Q_t) \cdot (\Lambda) := \sum_{\bk \in [r_1]\times[r_2] \cdots \times[r_t]} \Lambda(\bk) Q_1(\cdot, k_1) \otimes Q_2(\cdot, k_2) \cdots \otimes Q_t(\cdot, k_t), \label{eq:tucker_decomp}
\end{align}
and depicted in Figure \ref{fig:decomposition}. The higher order SVD (HOSVD) specifies a unique Tucker decomposition in which the factor matrices $Q_1 \dots Q_t$ are orthonormal and correspond to the left singular vectors of the unfolded tensor along each mode. Furthermore, the slices of the core tensor $\Lambda$ along each mode are mutually orthogonal with respect to entrywise multiplication, and the Frobenius norm of the slices of the core tensor are ordered decreasingly and correspond to the singular values of the matrix resulting from unfolding the tensor along each mode \cite{de2000multilinear}. The direct relationship between the HOSVD and the SVD of each unfolded tensor provides a direct method to compute the HOSVD and thus the multilinear rank from complete observation of a tensor. 

If the CP-rank is $r$, the Tucker-rank is bounded above by $(r,r,\dots r)$ by constructing a superdiagonal core tensor. If the Tucker rank is $(r_1, r_2, \dots r_t)$, the CP-rank is bounded by the number of nonzero entries in the core tensor, which is at most $r_1 r_2 \cdots r_t / (\max_{\ell} r_{\ell})$ \cite{de2000multilinear}. While the latent factors of the HOSVD are orthogonal, the latent factors corresponding to the minimal CP-decomposition may not be orthogonal.
%
For simplicity of presentation, we will first consider a limited setting where there exists a decomposition of the tensor into the sum of orthogonal rank-1 tensors. This is equivalent to enforcing that the core tensor $\Lambda$ associated to the Tucker decomposition is superdiagonal, or equivalently enforcing that the latent factors in the minimal CP-decomposition are orthogonal. There does not always exist such an orthogonal CP-decomposition, however this class still includes all rank 1 tensors which encompasses the class of instances used to construct the hardness conjecture in \cite{BarakMoitra16}. Our results also extend beyond to general tensors as well, though the presentation is simpler in the orthogonal setting.

As some of our matrix and tensor variables will have subscripts, we will index entries in the matrix or tensor by arguments in parentheses, i.e. $Q_{\ell}(a,b)$ denoting the entry in matrix $Q_{\ell}$ associated to index $(a,b)$. We will use bold font $\bi = (i_1, i_2, \dots i_t)$ to denote the index vector of a tensor and $i_{\ell}$ to denote the $\ell$-th coordinate of vector $\bi$. For data vectors, we use parentheses notation to access entries in the vector. We use $e_a$ to denote the standard basis vector with 1 at coordinate $a$ and zero elsewhere, and we use ${\bf 1}$ to denote the all ones vector. We denote the set $[n] = \{1, 2, \dots n\}$. $\Delta(\mathcal{S})$ denotes the probability simplex over $\mathcal{S}$. For $y \neq z \in [t]^2$, let $\cI_{yz}(a,b) = \{\bi \in [n_1]\times \dots \times [n_t] ~\text{s.t.}~ i_y = a, i_z = b\}$ denotes the set of indices $\bi$ such that the $y$-th coordinate is equal to $a$ and the $z$-th coordinate is equal to $b$. Similarly define $\cJ_{yz}(a,b) = \{\bk \in [r_1]\times \dots \times [r_t] ~\text{s.t.}~ k_y = a, k_z = b\}$.
\section{Key Intuition in a Simple Setting} \label{sec:intuition}

Consider a simple setting for a 3-order tensor that has low orthogonal CP-rank $r$, i.e. 
\begin{align}
T = \sum_{k= 1}^r \lambda_k Q_1(\cdot, k) \otimes Q_2(\cdot, k) \otimes Q_3(\cdot, k),
\end{align}
where $\otimes$ denotes an outer product and the columns of the latent factor matrices $Q_1 \in \Reals^{n_1 \times r}$, $Q_2 \in \Reals^{n_2 \times r}$, and $Q_3 \in \Reals^{n_3 \times r}$ are orthonormal. This is equivalent to assuming a Tucker decomposition consisting of the latent factor matrices $Q_1, Q_2,$ and $Q_3$, with a superdiagonal core tensor $\Lambda$ having $(\lambda_1, \lambda_2, \dots \lambda_r)$ on the superdiagonal.
Suppose that you were given additional information that for each $k \in [r]$ and $\ell \in [3]$, $|\langle Q_{\ell}(\cdot, k), \frac{1}{n_3} {\bf 1} \rangle| > \mu > 0$. Let us construct matrix $M_{12}^{obs}$ by averaging observed entries along the 3rd mode,
\begin{align} \label{eq:M12}
M_{12}^{obs}(a,b) = \frac{\sum_{i =1}^{n_3} T^{obs}(a,b,i) \Ind{(a,b,i) \in \Omega}}{\sum_{i =1}^{n_3} \Ind{(a,b,i) \in \Omega}}.
\end{align}
The expectation of $M_{12}^{obs}$ with respect to the randomness in the sampling pattern and observation noise is a rank $r$ matrix that has the same latent factors as the underlying tensor, i.e. conditioned on $(a,b)$ being observed, $\E{M_{12}^{obs}(a,b)} = e_a^T Q_1 \tilde{\Lambda} Q_2^T e_b$ where $\tilde{\Lambda}$ is diagonal with $\tilde{\Lambda}_{kk} = \lambda_k \langle Q_3(\cdot, k), \frac{1}{n_3} {\bf 1} \rangle$. 
The key observation is that the matrix $\E{M_{12}^{obs}(a,b)}$ and tensor $T$ share the same latent factor matrices $Q_1$ and $Q_2$. 
In Section \ref{sec:proof_sketch}, we show that this insight extends to the general context of a $t$-order tensor.
Repeating this construction along other modes would result in matrices whose expected values share the latent factor matrix $Q_3$ with the tensor.
A natural approach is to use these constructed matrices to estimate the latent factors of the original tensor along each mode. This insight is also used by \cite{kolda2015symmetric} in the context of tensor factorization with a fully observed tensor.

Denote the sparsity pattern of matrix $M_{12}^{obs}$ with $\tilde{\Omega}_{12}$, where $(a,b) \in \tilde{\Omega}_{12}$ iff there exists at least one $i \in [n_3]$ such that $(a,b,i) \in \Omega$, i.e. there is an observed tensor datapoint involving both coordinates $a$ and $b$. Due to the Bernoulli sampling model, $\Ind{(a,b) \in \tilde{\Omega}_{12}}$ is independent across all $(a,b) \in [n_1] \times [n_2]$, and $\Prob((a,b) \in \tilde{\Omega}_{12}) = 1 - (1-p)^{n_3} =: \tp \approx p n_3$, where the approximation holds if $p n_3 = o(1)$. As a result, the density of observations in $M_{12}^{obs}$ is $\tilde{p} = \Theta(\min(1, p n_3))$, which is significantly more dense than the original tensor dataset. 

In this simple setting where  $|\langle Q_{\ell}(\cdot, k), \frac{1}{n_3} {\bf 1} \rangle| > \mu$, we have reduced the task of estimating the latent factor matrices of a sparsely observed tensor $T$ to the easier task of estimating the latent factors of a not-as-sparsely observed matrix $\E{M_{12}^{obs}}$, where the data is also generated from a Bernoulli sampling model. As matrix estimation is very well understood, we can then apply methods from matrix estimation to learn the latent factors of $\E{M_{12}^{obs}}$. The non-zero singular values of $\E{M_{12}^{obs}}$ are $\lambda_k \langle Q_3(\cdot, k), \frac{1}{n_3} {\bf 1} \rangle$, which has magnitude bounded below by $|\lambda_k| \mu$ by assumption from the additional information. If this inner product were equal to zero for any value of $k$, it would imply that the rank of $\E{M_{12}^{obs}}$ is strictly smaller than the $r$ such that we would not be able to recover the full desired latent factor matrices for $T$ by estimating $\E{M_{12}^{obs}}$.

\begin{figure} 
	\centering
	\includegraphics[width=5in]{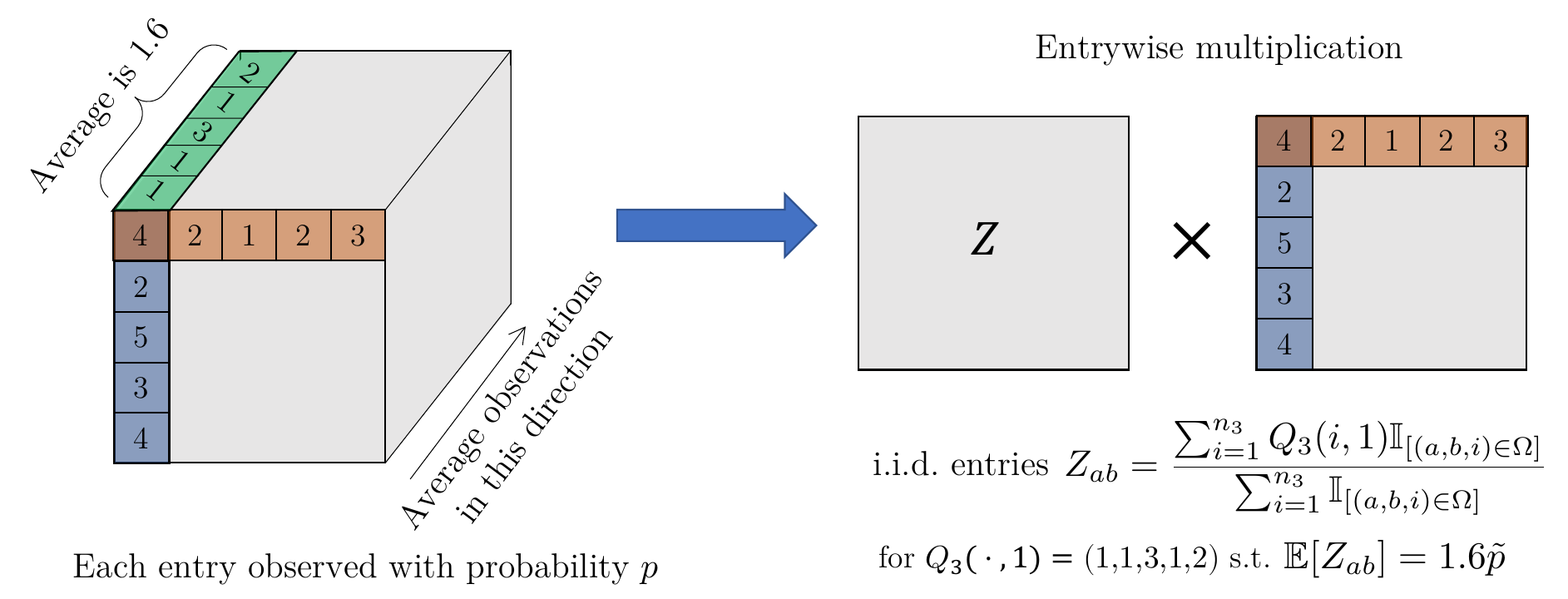}
	\caption{For a rank 1 tensor with no additive noise, averaging out entries in the third mode results in a matrix whose expected value is a rank 1 matrix with the same latent factors as the original tensor.} \label{fig:rank1ex}
\end{figure}

The distribution of $M_{12}^{obs}(a,b)$ depends not only on the original additive noise model of the tensor, but also the sampling process and the latent factors $Q_3$. For example the simplest rank-1 setting with exact observation of the tensor entries is depicted in Figure \ref{fig:rank1ex}. The observation $M_{12}^{obs}(a,b)$ takes value $\lambda Q_1(a,1) Q_2(b,1) Z_{ab}$, where
\begin{align}
Z_{ab} = \frac{\sum_{i =1}^{n_3} Q_3(i,1) \Ind{(a,b,i) \in \Omega}}{\sum_{i =1}^{n_3} \Ind{(a,b,i) \in \Omega}}.
\end{align}
This highlights that the noise is in fact a multiplicative factor $Z_{ab}$, where $\E{Z_{ab}} = \langle Q_3(\cdot, k), \frac{1}{n_3} {\bf 1} \rangle$ and the independence in the original tensor observations also implies $Z_{ab}$ are independent across indices $(a,b)$. Incoherence style assumptions on the latent factor guarantee that this multiplicative noise factor is not ill-behaved. While any matrix estimation algorithm could plausibly be used on $M_{12}^{obs}$, the theoretical results require analyses that are robust to more general noise models beyond the commonly assumed Gaussian additive noise model. When the data entries are bounded, we will use the approach in \cite{song2016blind, BorgsChayesLeeShah17, shah2019iterative} as they provide statistical guarantees for general mean zero bounded noise. 

Let's consider the impact on the sample complexity when the dimensions are equal, i.e. $n_1 = n_2 = \cdots n_t = n$. In contrast to approaches that unfold the tensor to a matrix, our algorithm collapses the tensor such that the constructed matrix has only $O(n^2)$ entries rather than $O(n^t)$ entries. Since matrix estimation algorithms require sample complexity scaling linearly in the maximum dimension of the matrix, this can be high for flattenings of the tensor to a matrix, in fact the maximum matrix dimension will always be at least $O(n^{t/2})$. By collapsing the tensor via averaging out other modes, the maximum matrix dimension of $M_{12}^{obs}$ is only linear in $n$. As a result estimating $\E{M_{12}^{obs}}$ only requires roughly linear in $n$ observations, which is significantly lower than the $O(n^{t/2})$ sample complexity common in tensor estimation. 

To summarize, the key insight is that while there may not be sufficient observations to directly perform inference over the tensor $T^{obs}$ due to high noise or sparsity, the constructed matrix $M_{12}^{obs}$ can be significantly less noisy and sparse due to aggregating many datapoints. As low rank matrix completion algorithms only requires observing roughly linear number of entries in order to recover the underlying matrix, the goal is to show that we could recover the underlying tensor $T$ when the number of observations is roughly linear in the dimension $n$, by applying matrix completion methods to appropriately constructed $n \times n$ matrices. 

\subsection{Sufficient Side Information}

While we sketched the intuition for the simplest setting with low orthogonal CP-rank and the condition that $|\langle Q_{\ell}(\cdot, k), \frac{1}{n_{\ell}} {\bf 1} \rangle| > \mu > 0$, this approach can be extended. The key property we needed was knowledge of some weight vector $W_{\ell} \in\Reals^{n_{\ell}}$ such that $|\langle Q_{\ell}(\cdot, k), \frac{1}{n_{\ell}} W_{\ell} \rangle| > \mu > 0$. Given such a weight vector, instead of computing $M_{12}^{obs}$ according to a uniform averaging of observed entries as in \eqref{eq:M12}, we simply compute a modified average using the given weight vector according to,
\begin{align} \label{eq:M12_wt}
M_{12}^{obs}(a,b) = \frac{\sum_{i =1}^{n_3} T^{obs}(a,b,i) W_3(i) \Ind{(a,b,i) \in \Omega}}{\sum_{i =1}^{n_3} \Ind{(a,b,i) \in \Omega}}.
\end{align}
The expectation of $M_{12}^{obs}$ would still share the same latent factor matrices as the tensor $T$, with $\E{M_{12}^{obs}(a,b)} = e_a ^T Q_1 \tilde{\Lambda} Q_2^T e_b$ for $(a,b) \in \tilde{\Omega}$ where $\tilde{\Lambda}$ is diagonal with $\tilde{\Lambda}_{kk} = \lambda_k \langle Q_3(\cdot, k), \frac{1}{n_3} W_{\ell} \rangle$, which has magnitude bounded away from zero by assumption on $W_{\ell}$. As a result, we could apply the same approach of estimating the latent factor matrices of the tensor using the more densely populated matrix $M_{12}^{obs}(a,b)$. In the same way that we will require incoherence style regularity conditions on $Q_{\ell}$, under the weighted model, we additionally require regularity conditions on the weighted latent factor matrices $\text{diag}(W_{\ell}) Q_{\ell}$.

When does such side information in the form of these specified weight vectors exist? And when would it be reasonable to assume knowledge of such side information? 
The condition that $|\langle Q_{\ell}(\cdot, k), \frac{1}{n_{\ell}} W_{\ell} \rangle| > \mu > 0$ for all $\ell,k$ simply imposes that the properly scaled weight vector $W_{\ell}$ must not be nearly orthogonal to any of the latent factors $Q_{\ell}(\cdot, k)$. In comparison to other common models of side information, this is significantly weaker than assuming full knowledge of the column space of $Q_{\ell}$. The requirement that $W_{\ell}$ is not directly aligned to the latent factors is mild, for example it would be satisfied by any linear combination of the latent factors with coefficients bounded away from zero, e.g. choosing a weight vector according to $\frac{1}{n_{\ell}} W_{\ell} = \sum_{k =1}^r \mu_{\ell} Q_{\ell}(\cdot, k)$ would satisfy $\langle Q_{\ell}(\cdot, k), \frac{1}{n_{\ell}} W_{\ell} \rangle = \mu_{\ell}$. If the latent factor matrices $Q_{\ell}$ satisfy incoherence-like properties, the vector $W_{\ell}$ would also satisfy similar regularity conditions.
As a result, given observed features of the indices, it is relatively mild to assume that one can construct a vector that satisfies the desired properties for the side information.


When the tensor does not have low orthogonal CP-rank, i.e. the core tensor corresponding to the HOSVD is not superdiagonal, the equivalent condition for sufficient side information is slightly more involved. We discuss the formal conditions for general tensors in Section \ref{sec:thm}. Also, one may ask whether we could simply use the observed entries to construct a weight vector $W_{\ell}$, in which case one would not need to be provided the weight vector in advance as auxiliary side information. Although this could be possible in special scenarios, a case study of the set of hard instances used in \cite{BarakMoitra16} shows that this is not possible in fullest generality, which we discuss in section \ref{sec:without_side_info}.

\section{Algorithm}

Given sparse noisy observations of the tensor $T^{obs}$, and given side information in the form of a weight vector $W_{\ell} \in \Reals^{n_{\ell}}$ for each mode $\ell \in [t]$, the main approach is to transform the tensor estimation problem into a matrix problem where the latent factors of the matrix are the same as the latent factors of the tensor. Once we learn the latent factors of the tensor, we use the estimates to reconstruct the original tensor. While one could use different existing matrix algorithms for estimating the latent factor matrices given the constructed matrices, we choose to use a variant of the iterative collaborative filtering algorithm from \cite{borgs2021iterative} due to its ability to handle general bounded noise models.


Our algorithm uses a subroutine from the iterative collaborative filtering algorithm in \cite{borgs2021iterative} to compute distances $\hat{d}_y(a,b)$ that approximate $\|Q_y^T (e_a - e_b)\|_{W}$ for an appropriate $W$. Subsequently it uses those distance estimates to compute nearest neighbor estimates over the tensor dataset. For simpler notation, we assume that $n_1 = n_2 = \dots n_t = n$ and the density of observations $p = n^{-(t-1) + \kappa}$ for $\kappa > 0$. The analysis can be modified to extend to tensors with uneven dimensions as long as the dimensions scale according to the same order. An additional benefit of using a nearest neighbor style algorithm is that the analysis leads to bounds on the maximum entrywise error of the final estimate as opposed to aggregate mean squared error bounds typical in the literature.

\subsection{Formal Algorithm Statement}\label{sec:alg_formal}

To facilitate cleaner analysis, we assume access to three fresh samples of the dataset, $T_1^{obs}$, $T_2^{obs}$, and $T_3^{obs}$ associated to observation sets $\Omega_1, \Omega_2, \Omega_3$ respectively (this is not necessary and can be instead handled by sample splitting as illustrated in \cite{borgs2021iterative}). Each dataset is used for different part of the algorithm to remove correlation across computations in different steps. $T_1^{obs}$ is used to construct $M^{obs}_{yz}$ in \eqref{eq:M_obs}; $T_2^{obs}$ is used in the distance computation in \eqref{eq:dist}, and $T_3^{obs}$ will be used for the final estimates in \eqref{eq:nn_est}. 

\medskip \noindent
{\bf Phase 1 (Estimating distances):}
For each mode $y \in [t]$, choose some $z \neq y$ and follow Steps 1-3 to estimate distances $\hat{d}_y(a,b)$ for each pair of $(a,b) \in [n_y]^2$ using the datasets $T_1^{obs}$ and $T_2^{obs}$.

\medskip \noindent
{\em Step 1: Construct $M^{obs}_{yz}$ and associated data graph.} Construct $M^{obs}_{yz}$ using data in $T_1^{obs}$ according to 
\begin{align}
	M_{yz}^{obs}(a,b) &=  \frac{\sum_{\bi \in \cI_{yz}(a,b)} T_1^{obs}(\bi) \prod_{\ell \in [t] \setminus\{y,z\}} W_{\ell}(i_{\ell})}{ | \Omega \cap \cI_{yz}(a,b)| }, \label{eq:M_obs}
\end{align}
where recall that $\cI_{yz}(a,b) = \{\bi \in [n_1]\times \dots \times [n_t] ~\text{s.t.}~ i_y = a, i_z = b\}$, and $\bi = (i_1, i_2, \dots i_t)$. Let $\tilde{\Omega}_{yz}$ denote the index set of nonzero entries in $M^{obs}_{yz}$. Let $\cG_{yz}$ denote the bipartite graph with vertex sets $[n_y]$ and $[n_z]$ and edge set $\cE_{yz}$. The pair $(a,b) \in [n_y] \times [n_z]$ is an edge in $\cE_{yz}$ iff $(a,b) \in \tilde{\Omega}_{yz}$.

\medskip \noindent
{\em Step 2: Construct BFS trees and associated statistics.} For each $a \in [n_y]$, construct a breadth first search tree rooted at vertex $a$ using edges in $\cG_{yz}$. Let $\cS_{a,s}$ denote the set of vertices in $\cG_{yz}$ that are at distance $s$ from root vertex $a$. Let $\cB_{a,s}$ denote the set of vertices in $\cG_{yz}$ that are at distance at most $s$ from $a$, i.e. $\cB_{a,s} = \cup_{h=0}^s \cS_{a,h}$. Let $\text{path}(a,i)$ denote the set of edges along the shortest path from vertex $a$ to $i$. Define neighborhood vectors 
\begin{align}
N_{a,s}(i) = \Ind{i \in \cS_{a,s}} \prod_{(u,v) \in \text{path}(a,i)} M^{obs}_{yz}(u,v).
\end{align}
Denote normalized neighborhood vectors as $\tN_{a,s} = N_{a,s}/|\cS_{a,s}|$. Choose the depth $s = \lceil \ln(n)/\ln(pn^{t-1}) \rceil$. 

\medskip \noindent
{\em Step 3: Distance computation.} For each $a \neq b \in [n]^2$, compute distance estimates $\hat{d}_y(a,b)$ via
\begin{align}
	\hat{d}_y(a,b) &= D(a,a) + D(b,b) - D(a,b) - D(b,a) \label{eq:dist} \\ 
	D(u,v) &= \frac{1}{p n^{t-2}} \sum_{i,j} \tN_{u,s}(i) \tN_{v,s+1}(j) \sum_{\bh \in \cI(i,j)} T_2^{obs}(\bh)  \prod_{\ell \in [t] \setminus\{y,z\}} W_{\ell}(h_{\ell})\label{eq:dist_D}
\end{align}
where $\cI(i,j) = \cI_{yz}(i,j)$ for even values of $s$ and $\cI(i,j) = \cI_{yz}(j,i)$ for odd values of $s$.

\medskip \noindent
{\em Step 4: Latent subspace computation (optional).} While the distance estimates in $\hat{d}_y$ will be sufficient to compute the nearest neighbor estimator described below in Phase 2, if one desires to directly obtain an estimate for the latent subspaces, one could compute the singular value decomposition of the symmetric matrix corresponding to the distance estimates $\hat{d}_y$, and let $\hat{Q}_y$ denote the eigenvectors corresponding to the top $r_y$ eigenvalues of the estimated distance matrix. $\hat{Q}_y$ serves as an $r_y$-dimensional approximation for the latent subspace along the $y$-th mode of the tensor. 

\medskip \noindent
{\bf Phase 2 (Nearest neighbor averaging):} Given the distance estimates $\hat{d}_y(a,b)$ for all $y \in [t]$ and $(a,b) \in [n_y]^2$, estimate the tensor using nearest neighbor averaging, 
\begin{align}
	\hat{T}(\bi) = \frac{\sum_{\bi' \in \Omega_3} T_3^{obs}(\bi') K(\bi, \bi')}{\sum_{\bi' \in \Omega_3} K(\bi, \bi')}
	~~~\text{ for }~~~
	K(\bi,\bi') = \prod_{\ell \in [t]} \Ind{d_{\ell}(i_{\ell}, i'_{\ell}) \leq \eta}. \label{eq:nn_est}
\end{align}
where we use a simple threshold kernel with parameter $\eta = \Theta\left(\max(n^{-\kappa/2}, n^{-2(\kappa+1)/(t+2)})\right)$.

Let $m = |\Omega|$ denote the number of observations in the tensor, and let $m' = \max_{y,z} |\tilde{\Omega}_{yz}|$ denote the maximum number of observations in the matrices that are constructed from averaging out $t-2$ modes of the tensor, where the maximium is taken over $y,z$.
A naive upper bound on the computational complexity of the algorithm is $t(m + n m' + n m') + n^t m$, where the terms are from constructing the matrix $M_{yz}^{obs}$, for constructing the BFS trees associated to $\cG$, computing pairwise distances, and calculating the nearest neighbor estimator. One could improve the computational complexity of the algorithm by choosing only to compute pairwise distances between a subset of vertex pairs, and clustering the vertices. For an appropriate choice of number of clusters, this would achieve the same theoretical guarantees while having significantly reduced computation.

\subsection{Variations of the Algorithm}\label{sec:alg_variant}

We presented a specific algorithm above to illustrate how to use the given side information to simplify tensor completion. However, the basic approach of using the side information to reduce the tensor to a matrix along each mode can also lead to other natural algorithms arising from other matrix estimation algorithms, as presented below.

\medskip \noindent
{\bf Phase 1 (Estimate latent subspaces):} For each mode $y \in [t]$ and some $z \neq y$, construct the matrix $M^{obs}_{yz}$ according to equation \eqref{eq:M_obs}. Use any low rank matrix estimation algorithm to estimate $\E{M^{obs}_{yz}}$, denoting the output as $\hat{M}_{yz}$. Compute the singular value decomposition of $\hat{M}_{yz}$, and let $\hat{Q}_y$ denote the left singular vectors corresponding to the top $r_y$ largest singular values of $\hat{M}_{yz}$. $\hat{Q}_y$ is as an $r_y$-dimensional approximation for the latent subspace along the $y$-th mode of the tensor. 

\medskip \noindent
{\bf Phase 2 (Least squares minimization):} Using the matrices $\hat{Q}_y$ as an approximation for the latent factor matrices $Q_y$, the tensor can be approximated by a multilinear multiplication of the latent factor matrices $\hat{Q}_1, \hat{Q}_2, \dots \hat{Q}_t$ with a $r_1 \times r_2 \times \cdots r_t$ core tensor. Compute $\hat{T}$ by solving the following unconstrained convex program minimizing the squared error
\begin{align}
\text{minimize} \quad&\sum_{\bi \in \Omega} (T^{obs}(\bi) - \hat{T}(\bi))^2 \\
\text{s.t. }\quad &\hat{T} = (\hat{Q}_1 \otimes \cdots \hat{Q}_t) \cdot (\hat{\Lambda}) \\
& \hat{\Lambda} \in \Reals^{r_1 \times r_2 \times \cdots r_t}
\end{align}


Empirically, we find that both proposed variants of Phase 1 are not computationally costly (unless the matrix estimation subroutine chosen is itself costly), but the nearest neighbor averaging in Phase 2 can be slow. The least squares minimization variant of Phase 2 is both faster and seems to perform slightly better in practice, as it does not require tuning of hyperparameters, whereas nearest neighbor averaging is sensitive to the choice of the averaging threshold. The nearest neighbor averaging variant is still useful to present though as it facilitates an easier analysis for showing our significant gains in sample complexity. 
\section{Theoretical Guarantees} \label{sec:thm}

We will present theoretical guarantees for the algorithm presented in Section \ref{sec:alg_formal}, which builds upon the iterative collaborative filtering style matrix estimation algorithm; as a result the model assumptions and analysis are similar to \cite{borgs2021iterative}. The first two are standard assumptions on uniform sampling and mean zero bounded observation noise.

\begin{assumption}[Sampling Model] \label{ass:sampling}
The set of observations is draw from a uniform Bernoulli$(p)$ sampling model, i.e. each entry $\bi \in [n_1] \times [n_2] \times \cdot [n_t]$ is observed independently with probability $p$.
\end{assumption}

\begin{assumption}[Observation Noise]
Each observation is perturbed by independent mean zero additive noise, i.e. $T^{obs}(\bi) = T(\bi) + E(\bi)$ where $E(\bi)$ are mean zero and independent across indices. Furthermore we assume boundedness such that $|T^{obs}(\bi)| \leq 1$.
\end{assumption}

The boundedness on $T^{obs}(\bi)$ implies that the noise terms $E(\bi)$ are also bounded and and also allow for heteroskedastic noise. The next assumption is in lieu of the standard incoherence assumption, which imposes regularity. We assume a latent variable model, in which the coordinates of each mode $\ell$ are associated to latent variables that are drawn from a population distribution, which for simplicity of notation is modeled as uniform over the unit interval. The population distribution can be easily extended to general distributions over a bounded set.

\begin{assumption}[Latent Variable Model]
For each mode $\ell \in [t]$, each coordinate $i_{\ell} \in [n_{\ell}]$ is associated to an i.i.d. sampled latent variable $x_{\ell}(i_{\ell}) \sim U[0,1]$. Each mode is associated to a set of $r$ bounded orthonormal functions $\{q_{\ell k}\}_{k \in [r]}$ with respect to $U[0,1]$, i.e. $|q_{\ell k}(x)| \leq B$ and 
\begin{align}
\E{q_{\ell k}(x) q_{\ell h}(x)} = \int_0^1 q_{\ell k}(x) q_{\ell h}(x) d x = \Ind{k=h}.
\end{align}
The ground truth tensor can be described by a bounded latent function $f$ of the latent variables with finite spectrum $r$, where $|f(\bx)| \leq 1$ and $f$ can be decomposed according to
\begin{align}
T(\bi) = f(x_1(i_1),x_2(i_2), \dots x_t(i_t)) = \sum_{\bk \in [r]^t} \Lambda(\bk) q_{1k_1}(x_1(i_1))q_{2k_2}(x_2(i_2)) \cdots q_{tk_t}(x_t(i_t)) \label{eq:tensor_latentf}
\end{align}
for some core tensor $\Lambda \in \Reals^{[r]^t}$.
\end{assumption}

The decomposition in \eqref{eq:tensor_latentf} implies that the tensor $T$ can be written according to a Tucker decomposition as in \eqref{eq:tucker_decomp} with latent factor matrices $Q_{\ell}(i,k) = q_{\ell k}(x_{\ell}(i))$ and core tensor $\Lambda(\bk)$ having dimension $r\times r\times \cdots r$. This implies the multilinear rank or Tucker rank is bounded above by $(r,r, \dots r)$. 
The multilinear rank of the tensor $T$ will be given by the multilinear rank of the core tensor $\Lambda$. As formally stated, we assume the latent variables are sampled from $U[0,1]$, but this can be relaxed to any general population distribution $P_{\ell}$ over a bounded set.

The latent variable model assumptions impose a distribution over the latent factors of the tensor such that in expectation over the randomness in the latent factors, they are orthogonal.
 The latent variables induce a distribution on the latent factors of the low rank decomposition via the functions $q_{\ell k}(x)$. 
For sufficiently large $n_1, n_2, \dots n_t$, the distribution of the sampled latent variables will be close to that of the population distribution, such that the finite averages over the sampled latent variables will converge to the population mean over the latent variable distribution. 
As a result, $Q_1 \dots Q_t$ are random latent factor matrices that have orthogonal columns in expectation with respect to the latent variable distribution, and as $n_1, n_2, \dots n_t \to \infty$ the columns of the sampled latent factor matrices of the tensor will be approximately orthogonal with high probability.

Along with the boundedness property, the generative model over the latent factors also guarantees incoherence style regularity conditions with high probability. Intuitively, incoherence attempts to formalize the concept that the signal is well spread amongst the coordinates and no single coordinate contains a critical component of the signal that is not exhibited by any other coordinate. When the latent factors are sampled from an underlying population distribution over a bounded set, then for sufficiently large $n_{\ell}$, there will be other coordinates that exhibit similar latent variables, and thus carry similar information content.
This assumption also helps to ensure that the ground truth tensor is not too empty. In particular, as the latent function is independent of the dimensions $n_1, \dots n_t$ and sparsity $p$, even if the latent function $f$ may take value zero for some subset of the latent feature space, it is still a constant measure with respect to the dimensions and sparsity parameters. As a result, the fraction of indices $\bi$ for which $\bi \in \Omega$ and $f(\bi) \neq 0$ will scale as $\Theta(p)$.

\begin{assumption}[Lipschitzness] \label{ass:lipschitz}
The latent function $f(\bx)$ is $L$-Lipschitz with respect to the 1-norm over the latent vector $\bx \in [0,1]^t$.
\end{assumption}

The Lipschitz assumption is used only in the analysis of the final estimate computed from nearest neighbor averaging. This assumption can be replaced by conditions on the induced distribution over $Q_{\ell}(i,\cdot)$ guaranteeing that for any index $i \in [n_{\ell}]$, there exists sufficiently many indices $j \in [n_{\ell}]$ such that $\|Q_{\ell}^T (e_i-e_j)\|_2$ is small, such that averaging over similar indices in the nearest neighbor estimator will converge.

\begin{assumption}[Side Information] \label{ass:side_info}
Side information is provided in the form of weight vectors $W_{\ell} \in [-1,1]^{n_{\ell}}$ such that their values are consistent with associated weight functions $w_{\ell}: [0,1] \to [-1,1]$ evaluated on the latent variables, i.e. $W_{\ell}(i) = w_{\ell}(x_{\ell}(i))$. Furthermore, for each pair $y \neq z \in [t]^2$, the matrix $\hat{\Lambda}_{yz} \in \Reals^{r \times r}$ defined by
\begin{align} \label{eq:hat_Lambda}
	\hat{\Lambda}(a,b) &:=  \sum_{\bk \in \cJ_{yz}(a,b)} \Lambda(\bk) \prod_{\ell \in [t] \setminus \{y,z\}} \frac{1}{n} \sum_{i \in [n]} q_{\ell k}(x_{\ell}(i)) w_{\ell}(x_{\ell}(i)).
\end{align}
is well conditioned with high probability with respect to the latent variables, i.e. the condition number of the matrix $\tilde{\Lambda}$ is bounded by a constant. Recall that $\cJ_{yz}(a,b) = \{\bk \in [r]^t ~\text{\normalfont s.t.}~ k_y = a, k_z = b\}$.
\end{assumption}

When the core tensor $\Lambda$ is superdiagonal, i.e. only having nonzero entries in indices $\bk = (k,k, \dots k)$, then the assumption on side information is equivalent to assuming that for each $\ell \in [t]$ and $k \in [r]$, $\frac{1}{n} \sum_{i \in [n]} q_{\ell k}(x_{\ell}(i)) w_{\ell}(x_{\ell}(i))$ is bounded away from zero. Assumption \ref{ass:side_info} generalizes the intuition presented in Section \ref{sec:intuition} beyond low CP-orthogonal rank tensors as $\Lambda$ does not need to be superdiagonal for this condition to be satisfied. 
However, Assumption \ref{ass:side_info} does require the multilinear rank of the core tensor $\Lambda$ to be balanced and equal to $(r,r, \dots r)$
in order for there to exist such a weight function such that $\tilde{\Lambda}_{yz}$ is full rank and well-conditioned for all $(y,z)$.

For simplicity of notation, we present our main theorem for $n_1 = n_2 = \cdots = n_t = n$, however the result extends to uneven dimensions as well as long as they scale proportionally together.

\begin{theorem} \label{thm:main}
Assume the data generating model for the observation tensor $T^{obs}$ satisfies Assumptions \ref{ass:sampling} to \ref{ass:lipschitz}, and assume the side information $\{W_{\ell}\}_{\ell \in [t]}$ satisfies Assumption \ref{ass:side_info}. Under sparse observations with $p = n^{-(t-1) + \kappa}$ for $\kappa > 0$, with probability $1 - \frac{8(1+o(1))}{n}$, the max entrywise error of the estimate output by our algorithm is bounded above by 
\[\max_{\bi \in [n]^t} |\hat{T}(\bi) - T(\bi)| =  O\Big(\max\Big(\frac{\log^{1/4}(n^2)}{n^{\min(\kappa,1)/4}}, \frac{\log^{1/(t+2)}(n^{t+1})}{n^{(\kappa+1)/(t+2)}}\Big)\Big).\]
It follows then that the mean squared error is also bounded above by
\[\mathbb{E}\Big[\frac{1}{n^t} \textstyle\sum_{\bi \in [n]^t} (\hat{T}(\bi) - T(\bi))^2\Big] = O\Big(\max\Big(\frac{\log^{1/2}(n^2)}{n^{\min(\kappa,1)/2}}, \frac{\log^{2/(t+2)}(n^{t+1})}{n^{2(\kappa+1)/(t+2)}}\Big)\Big).\]
\end{theorem}

Theorem \ref{thm:main} implies that given our model assumptions for ultra-sparse settings where the density $p = n^{-(t-1) + \kappa}$ for any arbitrarily small $\kappa > 0$, with high probability the max entrywise error of the estimate output by our algorithm goes to zero. Our result suggests that given appropriate side information, the estimation task is no harder for tensor estimation relative to matrix estimation, requiring only nearly linear sample complexity. This nearly linear sample complexity is a significant improvement from the best sample complexity of $O(n^{t/2})$ achieved by polynomial time algorithm, and is even better than the best achieved bound of $O(n^{3/2})$ of any statistical estimator. This form of side information only requires weak signal of the latent subspaces, and is significantly easier to satisfy than assuming full knowledge of the latent subspaces, as has been commonly assumed for previous empirical works studying tensor completion with side information.


The proof of the main theorem revolves around showing that the estimated distances $\hat{d}_y(a,b)$ concentrate around a function of the true distances with respect to the tensor latent factors, i.e. $\|A Q_y^T (e_a -e_b)\|_2^2$ for a well conditioned matrix $A$.
Given bounds on the estimated distances, we only need to bound the performance of the nearest neighbor estimator, which follows from the independence of observation noise, the assumption that there are sufficiently many nearby datapoints, and a simple bias variance tradeoff.

\subsection{Discussion} \label{sec:discussion}
Our results assume a latent variable model and constant rank as it builds upon the algorithm and techniques from \cite{BorgsChayesLeeShah17}. These assumptions replace the typical incoherence style conditions, as they guarantee that with high probability the latent factors are incoherent. This is reasonable for real-world settings with high dimensional data, as it essentially imposes regularity that the dimensions are associated to an underlying population distribution which satisfies regularity properties. The Lipschitz assumption is used to analyze the final nearest neighbor estimator, and is not necessary if we directly estimate the tensor from the latent factors. 

Simulations suggest that the alternate variations of the algorithm using other matrix estimation algorithms also attain similar guarantees, which give us reason to believe that the theoretical guarantees also likely extend. However, the analysis of the matrix estimation algorithms would need to handle heteroskedastic noise (or just arbitrary bounded mean zero noise), as the variance across sampling of entries in the other $t-2$ modes would depend on how widely the entries vary across the tensor. Additionally one would need guarantees on the row recovery of latent factors, which many of the previous work do not have. Recent work by \cite{chen2019inference} provides stronger control on recovery of latent factors, however it assumes a Gaussian noise model and would need to be extended to allow for arbitrary bounded noise that could result from aggregating entries sampled along other modes of the tensor.

The most restrictive assumption on the tensor is that the multilinear rank must be balanced, i.e. equal to $(r, r, \dots r)$ for some constant value of $r$. When the multilinear rank is not evenly balanced, the proposed algorithm will not be able to fully recover information about the latent subspaces, since the dimension of the column space of $\tilde{\Lambda}_{yz}$ for any choice of weight vectors may be strictly lower than $r_y $, which is equal to the dimension of the column space of $T_{(y)}$, the unfolded tensor along the $y$-th mode. The limitation of our result follows from the specific construction of the matrix $M_{yz}^{obs}$ by computing a weighted averaging along all other modes of the tensor except $y$ and $z$. In particular, the $i$-th column of $M_{yz}^{obs}$ consists of a weighted average of mode-$y$ fibers of the tensor, but specifically restricted to fibers corresponding to the $i$-th column of the slices of the tensor along modes $(y,z)$, weighted according to the same weight vector. As such, the dimension of the column space of the expected matrix $\E{M_{yz}^{obs}}$ is limited by the dimension of the column space of unfolding of the tensor along mode-$z$, which could be smaller if $r_z < r_y$. A possible extension of our idea that we leave for future exploration is to construct a different type of matrix $M^{obs}_y$ which consist of columns derived from weighted combinations of the mode-$y$ fibers, not limited to aligning fibers along slices of the tensor. The goal would be to construct $M^{obs}_y$ such that the column space of expected matrix $\E{M^{obs}_y}$ is equal to the column space of $T_{(y)}$. This seems plausible as the mode-$y$ fibers span the desired subspace. The weighted combinations should combine a significant number of the mode-$y$ fibers in order to guarantee that even when the dataset is only sparsely sampled, the density of observations in $M^{obs}_y$ is sufficiently large.

\subsection{In the Absence of Side Information} \label{sec:without_side_info}
The requirements of the side information for our result are weak and minimal in contrast to the assumptions made by other works in the literature, which either assume knowledge of the latent subspace \cite{bertsimastensor, zhou2017tensor, budzinskiy2020note, nimishakavi2018inductive,chen2019collective}, or similarity or kernel matrices that express relationship amongst the coordinates \cite{narita2012tensor, lamba2016incorporating}. We only require knowledge of a weight vector for which taking a weighted average along the specified mode does not zero out any component of the signal, which is far from requiring knowledge of the latent subspaces. 
One may ask whether it could be possible to construct the side information perhaps by sampling random weight vectors, or computing it as a function of the data itself. If this were possible, it would suggest an algorithm achieving nearly linear sample complexity even without side information. In general this is not possible without further conditions restricting the class of possible tensors. For example, consider the class of rank-1 tensors used to argue the hardness result from \cite{BarakMoitra16}. The latent factor vector $\theta \in \{-1, +1\}^{n}$ consists of randomly sampled entries where $\theta(i)$ for $i \in [n]$ is $-1$ with probability $1/2$ and $+1$ with probability $1/2$. As a result, for any choice of weight vector $W \in \Reals^n$ which is not too sparse, the expected value of $\langle \theta, W \rangle$ is zero, and furthermore with high probability, the typical vector $\theta$ will also result in a small inner product with the weight vector due to the symmetry in the generating distribution. As a result, the constructed matrices $M_{yz}^{obs}$ would have an expected value of zero, clearly not preserving the latent subspace. Any class of tensors with such symmetry would also exhibit similar properties; for example if we sampled the latent factor matrices by normalizing random matrices where each entry is sampled from an independent mean-zero Gaussian, with high probability the latent factors would be orthogonal to any fixed or randomly sampled weight vector.
However, one could argue that such perfect symmetry is in fact not common in real-world datasets. In section \ref{sec:experiments}, we benchmark our algorithm on both synthetic and real-world datasets, and we find that this assumption on side information is in fact satisfied for the real-world datasets derived from traffic and MRI measurements. We conjecture that in fact many real-world datasets are not perfectly symmetric, such that the naive all ones weight vector would perform well; this only requires that the sums of the latent factors are not too close to zero.

The ability to collect a limited amount of active samples could alternately replace the requirement for side information. If for each mode $\ell \in [t]$ we actively sampled an entire mode-$\ell$ fiber at random, then the sampled vector could be used as the weight vector $W_{\ell}$. The mode-$\ell$ fiber lies in the desired column space by definition, and by sampling a fiber at random, it is likely that the sampled vector would satisfy the required conditions. This suggests that given a budget of $t n$ active samples, we could produce a consistent estimate of the full tensor with nearly linear sample complexity, in contrast to the $n^{t/2}$ sample complexity without the active samples. This is consistent to the result in \cite{zhang2019}, which states that with an active sampling scheme, one can estimate the full tensor with $O(r t n)$ samples by directly estimating the latent subspaces from the actively sampled datapoints. Our result shows that limited ability to collect samples actively, can still be immensely beneficial even if the active samples are not sufficient themselves to fully recover the tensor.

\subsection{Proof Sketch} \label{sec:proof_sketch}

While we defer the formal proof to the appendix, we give the rough intuition and sketch of the proof here. The main property that the approach hinges on is that the column space of $\E{M_{yz}^{obs}}$ is the same as the column space of $T_{(y)}$, and furthermore the matrix $M_{yz}^{obs}$resulting from collapsing the tensor satisfies the desired regularity conditions in order to use a similar analysis as \cite{borgs2021iterative}.

First, we show that the expected value of the constructed matrix, with respect to the additive noise terms and the sampling distribution conditioned on the latent variables, indeed shares the same column space as the original tensor. 
\begin{align}
&\E{\left.M_{yz}^{obs}(a,b) ~\right|~ (a,b) \in \tilde{\Omega}, \{x_{\ell}(i)\} } \\
&= \E{\left.\frac{\sum_{\bi \in \cI_{yz}(a,b)} T_1^{obs}(\bi) \prod_{\ell \in [t] \setminus\{y,z\}} W_{\ell}(i_{\ell})}{ | \Omega \cap \cI_{yz}(a,b)| } ~\right|~ (a,b) \in \tilde{\Omega}, \{x_{\ell}(i)\} }\\
&= \sum_{\bi \in \cI_{yz}(a,b)} \E{\left.T(\bi) \prod_{\ell \in [t] \setminus\{y,z\}} W_{\ell}(i_{\ell}) ~\right|~ \{x_{\ell}(i)\} } \E{\left. \frac{\Ind{\bi \in \Omega}}{ | \Omega \cap \cI_{yz}(a,b)| } ~\right|~ (a,b) \in \tilde{\Omega}} 
\end{align}
\begin{align}
&= \sum_{\bi \in \cI_{yz}(a,b)} \sum_{\bk \in [r]^t} \Lambda(\bk) q_{y k_y}(x_y(a)) q_{z k_z}(x_z(b)) \prod_{\ell \in [t] \setminus\{y,z\}} q_{\ell k_{\ell}}(x_{\ell}(i_{\ell})) w_{\ell}(x_{\ell}(i_{\ell})) \frac{1}{n^{t-2}}\\
&= \sum_{j,h \in [r]^2} q_{y j}(x_y(a)) q_{z h}(x_z(b)) \sum_{\bk \in \cJ_{yz}(j,h)} \Lambda(\bk) \prod_{\ell \in [t] \setminus\{y,z\}} \frac{1}{n} \sum_{i=1}^n  q_{\ell k_{\ell}}(x_{\ell}(i)) w_{\ell}(x_{\ell}(i)) \\
&= e_a^T Q_y \hat{\Lambda} Q_z^T e_b
\end{align}
for matrix $\hat{\Lambda}$ as defined in \eqref{eq:hat_Lambda}.
If we take expectation with respect to the latent variables, then it follows that $\hLam$ concencentrates around $\tilde{\Lambda}_{yz}$, which is full rank by Assumption \ref{ass:side_info}. Let the SVD of $\hat{\Lambda}$ be denoted $\hat{U} \hat{\Sigma} \hat{V}^T$, where the diagonal entries of $\hSig$ are denoted as $\hsig_1 \dots \hsig_r$. Let $\tilde{Q}_y = Q_y \hU$ and $\tilde{Q}_z = Q_z \hV$ with the associated latent functions $\tilde{q}_{yk}, \tilde{q}_{zk}$ defined similarly. It follows then that $M_{yz}^{obs}$ is modeled by a latent variable model where the latent function is low rank and $L$-Lipshitz.
\begin{align} \label{eq:Ex_M_obs}
\E{M_{yz}^{obs}(a,b) ~|~ (a,b) \in \tilde{\Omega}} 
= e_a^T \tQ_y \hat{\Sigma} \tQ_z^T e_b
=: \tilde{f}(x_y(a),x_z(b)).
\end{align}
We can verify that the associated latent functions also exhibit orthonormality, e.g.
\begin{align}
\int_0^1 \tilde{q}_{y k}(x) \tilde{q}_{y h}(x) d x 
&= \int_0^1 \left(\sum_{i \in [r]} q_{y i}(x) \hU(i,k)\right) \left(\sum_{j \in [r]} q_{y j}(x) \hU(j,h)\right) d x \\
&= \sum_{i, j \in [r]^2} \hU(i,k) \hU(j,h) \int_0^1 q_{y i}(x) q_{y j}(x) d x
= \Ind{k=h}.
\end{align} 
Let $\tilde{B}$ be an upper bound on the latent functions $\tilde{q}_{\ell k}$ such that $|\tilde{q}_{\ell k}(x)| \leq \tilde{B}$. Due to boundedness of the functions $q_{\ell k}$, $\tilde{B}$ is at most $B r$. As a result, the latent function associated to $\tilde{f}(x_y(a),x_z(b))$ is rank $r$ with the decomposition
\begin{align}
 \tilde{f}(x,x') = \sum_{k=1}^r \hsig_k  \tilde{q}_{y k}(x) \tilde{q}_{y k}(x').
\end{align}
Furthermore we can verify that $\tilde{f}$ is also $L$-Lipschitz, inheriting the property from the latent function $f$. We use $f(x_y, x_z, \bx_{-yz}(\bi))$ to denote the latent function $f$ evaluated at the vector having values $x_y$ and $x_z$ in the $y$ and $z$ coordinates, and having values $x_{\ell}(i)$ for all other coordinates $\ell \in [t] \setminus \{y,z\}$. 
\begin{align}
|\tilde{f}(x_y,x_z) - \tilde{f}(x'_y,x'_z)| 
&= \left|\frac{1}{n^{t-2}} \sum_{\bi \in [n]^{t-2}} ( f(x_y, x_z, \bx_{-yz}(\bi)) -  f(x'_y, x'_z, \bx_{-yz}(\bi))) \prod_{\ell \in [t] \setminus\{y,z\}} W_{\ell}(i_{\ell}) \right| \nonumber \\
&\leq L |x_y - x'_y| + L|x_z - x'_z| \label{eq:tf_Lipschitz}
\end{align}
where the last inequality follows from Lipschitzness of $f$ as well as the boundedness assumption on $|W_{\ell}(i_{\ell})| \leq 1$.
This setup nearly satisfies the same data generating model that is used in \cite{borgs2021iterative}. A minor difference is that our model is asymmetric, however this leads to very little change in the analysis. The key difference in our model is that the distribution over $M_{yz}^{obs}(a,b)$ is more involved as it arises from the averaging of values over the other modes of the tensor. Furthermore we consider a wider range of densities $\tilde{p}$, from as sparse as $O(n^{-1+\kappa})$ to as dense as constant.

The proof relies on Lemma \ref{lem:dist_conc}, stated below, which shows that the estimated distances that our algorithm produces will approximate a function of the true distances with respect to the tensor latent factors. The first step is to argue that the sparsity of the observed local neighborhood graph grows sufficiently quickly, which follows from standard arguments for the properties of an Erdos-Renyi graph that results from the Bernoulli sampling model. The second step is a careful martingale concentration argument to show that $e_k ^T \tQ_y \tN_{a,j} \approx e_k^T \hSig^j \tQ_y e_a$. The third step is to use show that $D(a,a')$ concentrates around $\tN_{a,s}^T \tQ_y^T \hSig \tQ_z \tN_{a',s+1}$. Standard applications of concentration inequalities yield results that are too loose when $p$ is very sparse, i.e. $p = n^{-(t-1) + \kappa}$ for $\kappa \in (0,1)$. As a result we define a ``truncated'' modification of $D(a,a')$, which can be shown to concentrate well via a standard application of Bernstein's inequality, and we use the sparsity condition to argue that the truncated modification of $D(a,a')$ is in fact equal to $D(a,a')$ with high probability.
\begin{lemma} \label{lem:dist_conc}
Assume $p = n^{-(t-1) + \kappa}$ for $\kappa > 0$. Let $\tp = 1 - (1-p)^{n^{t-2}}$. With probability $1-\frac{6(1+o(1))}{tn}$, $\hat{d}_y(a,a')$ computed from \eqref{eq:dist} satisfies
\begin{align*}
	&\max_{a,a' \in [n]^2} |\hat{d}_y(a,a') - \|\hSig^{s+1} \tQ_y^T (e_a - e_{a'})\|_2^2| \leq \frac{5 r B \hsig_{\max}^{s+1} (1 + \hsig_{\max}^{s}) \log^{1/2}(trn^2) (1+o(1))}{(n\tp)^{1/2}}.
\end{align*}
\end{lemma}

Next Lemma \ref{lem:suff_many_nn} uses the Lipschitz and latent variable model assumptions to argue that there are sufficiently many nearest neighbors with respect to the distances defined by $\|\hSig^{s+1} \tQ_y^T (e_a - e_{a'})\|_2^2$. 
\begin{lemma} \label{lem:suff_many_nn}
For any $i \in [n]$, $\ell \in [t]$, $\eta' > 0$
\begin{align*}
\Prob\left(\sum_{i' \neq i \in [n]} \Ind{\|\hSig^{s+1} \tQ_y^T (e_a - e_{a'})\|_2^2 \leq \eta'} \leq (1-\delta) (n-1) \frac{\sqrt{\eta'}}{\hsig_{\max}^{s} L} \right)
&\leq \exp\left(-\frac{\delta^2 (n-1)\sqrt{\eta'}}{2\hsig_{\max}^{s} L}\right).
\end{align*}
\end{lemma}

The final steps of the proof of the main theorem follow from a straightforward analysis of the nearest neighbor averaging estimator given the estimated distances.

\section{Experiments} \label{sec:experiments}

We provide a sequence of experiments ranging from synthetic to real-world datasets in order to understand the performance of our proposed algorithm as compared against state-of-the-art tensor completion algorithms.
We benchmark the algorithms in the following four datasets.

\medskip \noindent
{\bf Synthetic:} We generate Gaussian matrices $U, V, W \in \Reals^{100 \times 10}$, with entries that are independently sampled from $N(1,1)$. We sample a diagonal core tensor $S \in [0,1]^{10\times10\times10}$, where each of the 10 diagonal entries is sampled from $U[0,1]$. Let $Z = (U \otimes V \otimes W) \cdot (S)$ be the multilinear multiplication of the core tensor $S$ with the latent factor matrices $U, V, W$. We then construct the ground truth tensor by normalizing $Z$ so that $|T(i,j,k)| \leq 1$, according to 
$T = Z/\max_{i,j,k} |Z(i,j,k)|$.
The ground truth tensor has rank 10 and dimensions $100 \times 100 \times 100$. For each observed location $(i,j,k) \in \Omega$, we add zero mean Gaussian observation noise with variance $\sigma^2$, for $\sigma = 0.1$. 

\medskip \noindent
{\bf MRI:} We construct a dataset in which the tensor $T$ is a scaled version of a volumetric MRI brain-scan. We used a volumetric MRI brain-scan dataset from the MIRIAD dataset, which contains brain scans of Alzheimer's sufferers and healthy elderly people \cite{malone2013miriad}. We used a single MRI scan from the dataset, which is represented as a $256 \times 256 \times 124$ tensor. We normalized the data such that the absolute value is bounded by 1. We do not add additional noise to the measurements, but study the performance under sparse sampling.

\medskip \noindent
{\bf Traffic:} We use the urban traffic speed dataset of Guangzhou, China from \cite{xinyu_chen_2018_1205229}. This is a dataset which consists of speed measurements from 214 anonymous road segments (consisting of urban expressways and arterials) from Guangzhou, China. The measurements were collected at 10-minute intervals from August 1, 2016 to September 30, 2016. The dataset is represented as a $214 \times 61 \times 144$ tensor, where the modes correspond to road segment, day, and time window. 
We normalized the data such that the absolute value is bounded by 1. We do not add additional noise to the measurements, but study the performance under sparse sampling.

\medskip \noindent
{\bf Random 3XOR:} We generate an instance of tensor completion which embeds a 3XOR formula, as described in \cite{BarakMoitra16}. We randomly sample a latent vector $a \in \{-1, +1\}^n$, where $a_i$ independently takes values $-1$ or $+1$ each with probability $1/2$. For each observed index $(i,j,k) \in \Omega$, 
\begin{align*}
T^{obs}(i,j,k) = 
\begin{cases}
a_i a_j a_k &\text{ with prob } 15/16, \\
-a_i a_j a_k &\text{ with prob } 1/16.
\end{cases}
\end{align*}
The expected tensor is rank 1, given by $T = \frac{7}{8} a \otimes a \otimes a$, such that $\E{T^{obs}(\bi)} = T(\bi)$ for all $\bi \in \Omega$. 

\medskip 
We will use the na\"ive all ones weight vector for the ``side information''. While this does not inherently encode extra information, we will see that the algorithm performs well for the Synthetic, MRI, and Traffic datasets. To verify the required condition on the weight vector, we took each of the ground truth tensors $T$, computed the true latent factor matrices, and computed the inner product between the column in the latent factor matrices and the all ones vector. For the Synthetic, MRI, and Traffic datasets, the minimum inner product across all modes and latent factors is $0.135, 0.0149,$ and $0.000992$, respectively. This supports our conjecture that real-world datasets are not perfectly symmetric, such that a na\"ive choice of weight vector may indeed suffice. In contrast, the inner product between the all ones vector and the latent factor in the 3XOR setting is as small as $9.71 \times 10^{-17}$. Consistent with the intuition from our analysis, our algorithm performs poorly for the random 3XOR dataset, as the latent factor is orthogonal to the weight vector.

\subsection{Implementation Details of Our Proposed Algorithm}

The key idea of our proposed algorithm is to reduce the tensor latent subspace estimation task to the simpler matrix latent subspace estimation task over a small matrix. We include implementations of three variations of our algorithm arising from different choices for Phases 1 and 2 of the algorithm.
\begin{itemize}
\item {\bf ICF\_NN} denotes the original algorithm as presented in Section \ref{sec:alg_formal}, which consists of estimating distances in Phase 1 followed by nearest neighbor averaging in Phase 2.
\item {\bf ICF\_LS} denotes the algorithm which uses Phase 1 from Section \ref{sec:alg_formal}, which estimates latent subspaces via the estimated distances as described in Step 4, followed by least squares minimization for Phase 2 as presented in Section \ref{sec:alg_variant}.
\item {\bf ALS\_LS} denotes the algorithm presented in Section \ref{sec:alg_variant}, using alternating least squares \cite{ZachariahSundinJanssonChatterjee12} for matrix estimation in Phase 1, followed by least squares minimization in Phase 2.
\end{itemize}
The depth parameter used in steps 2 and 3 of estimating distances in Phase 1 is chosen to be $s = 1$. The threshold for the nearest neighbor averaging in Phase 2 is chosen to be the 10th percentile of the estimated distances. We found that the algorithm performed well even without optimizing the hyperparameters; naturally the least squares variant (which has no hyperparameters) performed better than the nearest neighbor variant with the above fixed choice of hyperparameters. The rank hyperparameter for least squares minimization in Phase 2 is set to 10 along each mode for the Synthetic, MRI, and Traffic datasets, and set to 1 along each mode for the random 3XOR dataset.

\subsection{Other Algorithms for Comparison}

As a simple baseline, we compare with the na\"ive average, which simply fills in all missing entries with the average of all observed datapoints. When the data is extremely sparse, the benchmarked algorithms start to degrade significantly such that they perform worse than the na\"ive average. In addition, we compare against the following state of the art tensor completion algorithms. As all of these algorithms are iterative, the stopping criteria is met after a certain number of iterations (specified below) or when the difference between the iterates falls below a tolerance of $10^{-5}$.

\medskip \noindent
{\bf TenALS:} Tensor alternating least squares (TenALS) uses an alternating minimization based method to solve a non-convex least square problem, initializing with a tensor power method \cite{jain2014provable}. 
The authors prove that under incoherence assumptions, TenALS can provably recover a three-mode $n \times n \times n$ dimensional rank-$r$ tensor exactly from $O(n^{3/2} r^5 \log^4 n)$ randomly sampled entries \cite{jain2014provable}. 
We set the rank hyperparameter of TenALS to $10$ for the Synthetic, MRI, and Traffic datasets, and we set it to 1 for the random 3XOR dataset. We use the default of 10 initializations and 50 iterations. 


\medskip \noindent
{\bf LRTC:} Low rank tensor completion (LRTC) refers to a class of algorithms minimizing a variant of the tensor nuclear norm, which is a weighted average of the nuclear norms of all matrices unfolded along each mode \cite{liu2012tensor}. The authors do not provide statistical guarantees for their algorithms. We benchmark against SiLRTC, which uses block coordinate descent, and HaLRTC, which uses alternating direction method of multipliers (ADMM). 
For SiLRTC, we set the hyperparameters as: $\alpha = (\frac13,\frac13,\frac13)$ and $\beta =(1,1,1)$. Due to slow convergence, we set the number of maximum iterations to be $1000$. For HaLRTC, we set the weights as $\alpha = (\frac13,\frac13,\frac13)$. We adaptively increase the Lagrangian multiplier, setting the initial value to be $10^{-2}$, and increasing the multiplier by a factor of $1.1$ in each iteration. The upper bound for the multiplier is set to be $10^{10}$, and the number of maximum iterations is set to be $500$.

\medskip \noindent
{\bf TNN:} \cite{zhang2014novel} proposed a tensor nuclear norm (TNN) minimization algorithm which uses a definition of tensor nuclear norm constructed from the tensor-Singular Value Decomposition (t-SVD) proposed in \cite{kilmer2013thirdorder}. 
Under incoherence condtions, TNN can provably recover a three-mode $n \times n \times n$ dimensional rank-$r$ tensor exactly from $O(n^2 r \log n)$ randomly sampled entries \cite{zhang2016exact}. 
To minimize the tensor nuclear norm penalized objective function, they employ the general framework of ADMM. We set the initial value of the Lagrange multiplier to be $10^{-2}$, and increase it gradually by a factor of $1.1$ until it reaches $10^{10}$. The number of  maximum iterations is set to be $500$.

\subsection{Empirical Results}

In each of these experiments, we specifically pay attention to the performance at extreme levels of data sparsity, close to the threshold where the algorithms ``breaks'', i.e. performs worse than the na\"ive average baseline. As a result, we benchmark on values of $p \in [0,0.08]$. We evaluate the algorithm's performance with respect to the normalized mean squared error (MSE), defined as 
\begin{align*}
\text{Normalized MSE} := \frac{\sum_{i \in [n_1]}\sum_{j \in [n_2]} \sum_{k \in [n_3]} \left(\hat{T}(i,j,k) - T(i,j,k)\right)^2}{\sum_{i \in [n_1]}\sum_{j \in [n_2]} \sum_{k \in [n_3]} T(i,j,k)^2}
\end{align*}

\begin{figure}
     \centering
     \begin{subfigure}[b]{0.49\textwidth}
         \centering
         \includegraphics[width=\textwidth]{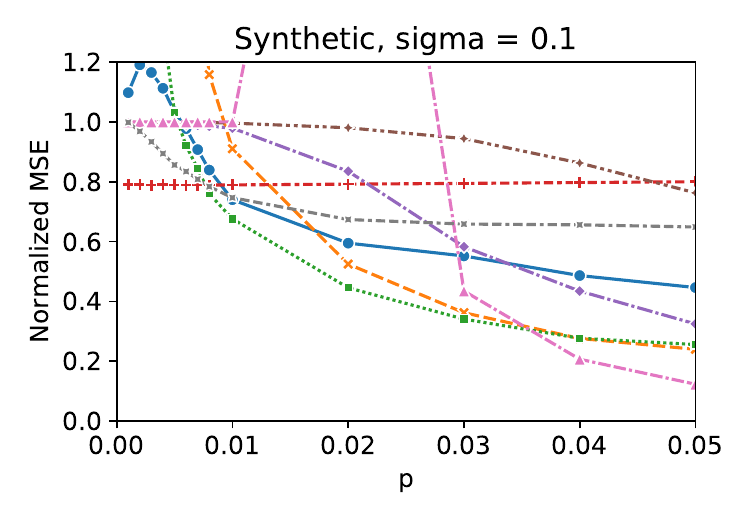}
     \end{subfigure}
     \begin{subfigure}[b]{0.49\textwidth}
         \centering
         \includegraphics[width=\textwidth]{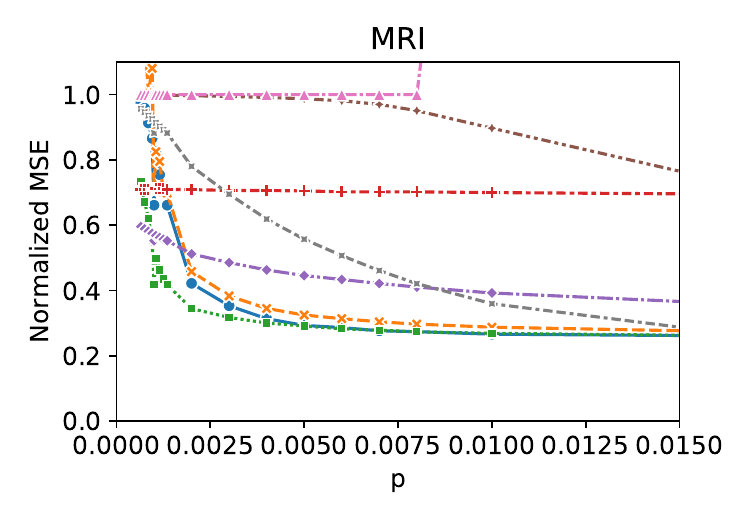}
     \end{subfigure} \\
     \begin{subfigure}[b]{0.49\textwidth}
         \centering
         \includegraphics[width=\textwidth]{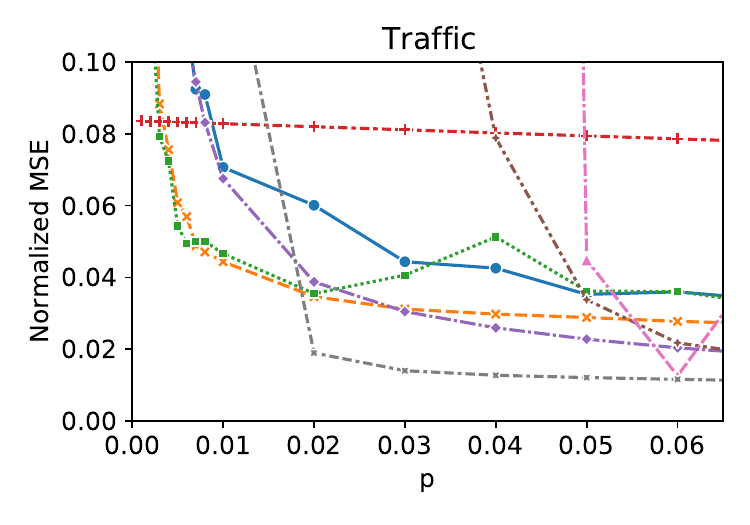}
     \end{subfigure}
     \begin{subfigure}[b]{0.49\textwidth}
         \centering
         \includegraphics[width=\textwidth]{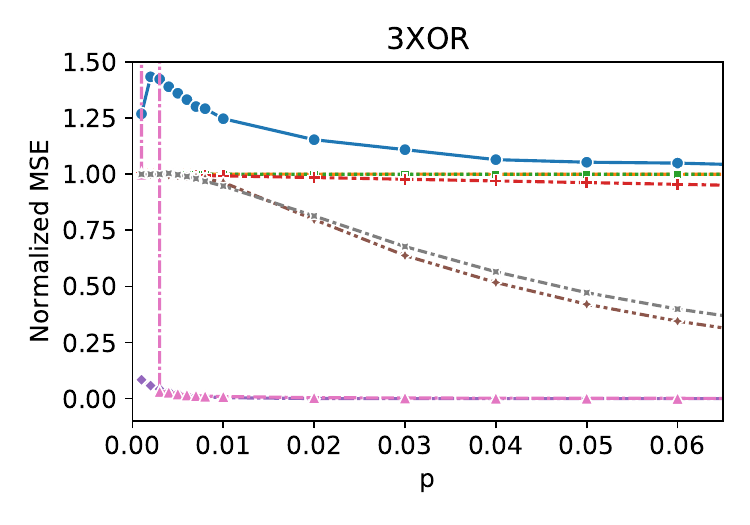}
     \end{subfigure} \\
     \begin{subfigure}[b]{0.55\textwidth}
         \centering
         \includegraphics[width=\textwidth]{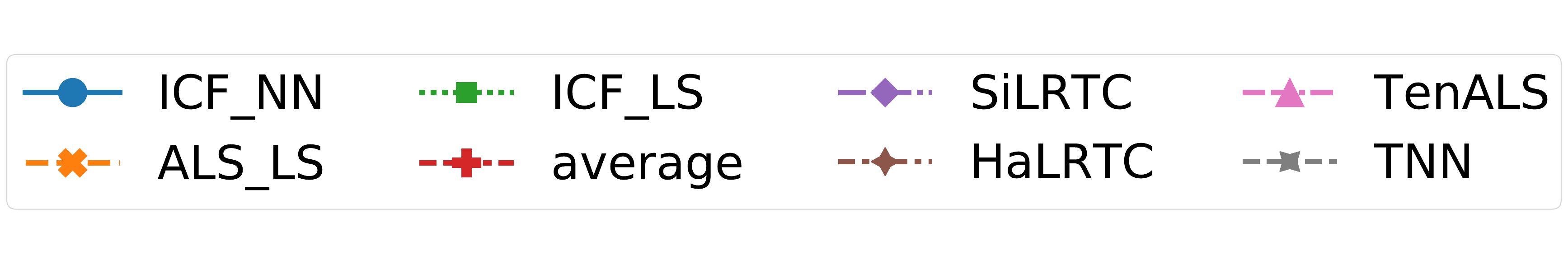}
     \end{subfigure}
        \caption{These plots show the normalized mean squared error attained by different tensor completion algorithms on a synthetic dataset, an MRI dataset, a traffic dataset, and a random 3XOR dataset. The scaling of the y and x axis are chosen to show the performance relative to the na\"ive average baseline on the extremely sparse settings near the limits of the algorithms' performance.} \label{fig:expt_results}
\end{figure}


Figure \ref{fig:expt_results} plots the normalized MSE achieved by the output of each algorithm as a function of the data sparsity $p$. The scale of the x-axis for each plot are set differently in order to highlight the results at the sparsest regimes near the lower threshold of performance. The scale of the y-axis for each plot are set to highlight the performance relative to the na\"ive average baseline, which is depicted by the red dashed line. Let us first focus on the results from the Synthetic, MRI, and Traffic datasets. Firstly, we notice that ICF\_LS consistently performs the best in the sparsest regimes relative to any other algorithm, followed by ALS\_LS and ICF\_NN. This suggests that our proposed algorithm in fact does do well in addressing data-poor settings. TenALS and HaLRTC perform very poorly in these sparse regimes, at times having MSE that is orders of magnitude higher than the average baseline (and thus not depicted in Figure \ref{fig:expt_results}). TNN and SiLRTC give reasonable performance, though it degrades more quickly than our proposed algorithms as the data becomes sparser. These experiments validate that our approach of reducing the tensor to a significantly smaller matrix to estimate the latent subspaces in fact does allow the algorithm to perform well at extreme levels of sparsity, even when the ``side information'' weight vector is na\"ively chosen to be the all ones vector. These examples also indicate that the side information requirements are indeed weak, and may not truly require ``additional information'' as long as the underlying model is not too symmetric.

In contrast, the random 3XOR dataset shows very different results. Our algorithms perform no better than the na\"ive average baseline, while TNN and HaLRTC show reasonably smooth degrading performance, and SiLRTC and TenALS perform extremely well even at low levels of sparsity. In fact, this example is chosen to illustrate when our algorithm will completely fail. The ground truth tensor is only rank-1, and the size of the tensor is not too large, i.e. $100 \times 100 \times 100$, so any brute force algorithm could perform very well. However, the entries in the latent factor are sampled independently from $\{-1,+1\}$, such that the latent factor is essentially orthogonal to the all ones vector. As our algorithm collapses the tensor along the third mode by summing according to the weight vector, this step unfortunately zeros out all the signal, resulting in a matrix that is essentially all zeros. As a result, our algorithms perform horribly poor on the random 3XOR dataset at all levels of sparsity $p$. Furthermore,it follows that without having access to secret knowledge of the latent factors, any attempt to construct a suitable weight vector for our algorithm will fail, as the symmetry in the distribution that the latent factors are sampled from will result in the latent factor being orthogonal in expectation to any chosen weight vector.


%
\begin{acks}
We gratefully acknowledge funding from the NSF under grants CCF-1948256 and CNS-1955997. Christina Lee Yu is supported by an Intel Rising Stars Faculty Award. Data used in the preparation of this article were obtained from the Urban Traffic Speed Dataset of Guangzhou, China published by Xinyu Chen, Yixian Chen and Zhaocheng He, and the MIRIAD database. The MIRIAD investigators did not participate in analysis or writing of this report. The MIRIAD dataset is made available through the support of the UK Alzheimer's Society (Grant RF116). The original data collection was funded through an unrestricted educational grant from GlaxoSmithKline (Grant 6GKC).
\end{acks}

\bibliographystyle{ACM-Reference-Format}
\bibliography{biblio}

\section{Appendix A: Proofs}
\subsection{Proof of Main Theorem}

The proof of the main theorem uses Lemmas \ref{lem:suff_many_nn} and \ref{lem:dist_conc}, whose proofs we defer to Sections \ref{sec:suff_many_nn} and \ref{sec:dist_conc}. Given the results in these lemmas, the remaining proof of the main theorem follows from a straightforward analysis of the nearest neighbor averaging estimator given the estimated distances.

\begin{new-proof}{Theorem \ref{thm:main}}
Recall that our final estimate is computed via a nearest neighbor,
\begin{align*}
\hat{T}(\bi) = \frac{\sum_{\bi'} T^{obs}_3(\bi') K(\bi, \bi')}{\sum_{\bi'} \Ind{\bi' \in \Omega_3} K(\bi, \bi')} ~\text{ where }~ K(\bi,\bi') = \prod_{\ell \in [t]} \Ind{\hat{d}_{\ell}(i_{\ell}, i'_{\ell}) \leq \eta}.
\end{align*}

We can decompose the entrywise error into a bias term and an observation noise term,
\begin{align*}
|\hat{T}(\bi) - T(\bi)|
= \frac{\sum_{\bi'} \Ind{\bi' \in \Omega_3} K(\bi, \bi') |T(\bi') - T(\bi)|}{\sum_{\bi'} \Ind{\bi' \in \Omega_3} K(\bi, \bi')} + \left|\frac{\sum_{\bi'} \Ind{\bi' \in \Omega_3} K(\bi, \bi') (T^{obs}_3(\bi') - T(\bi'))}{\sum_{\bi'} \Ind{\bi' \in \Omega_3} K(\bi, \bi')}\right|.
\end{align*}

Let us denote $\rho_y(h) = \sum_{\bk \in [r]^t} \Lambda(\bk) \Ind{k_y = h}$.
The bias term we can bound by 
\begin{align*}
|T(\bi') - T(\bi)|
&= \left|\sum_{\bk \in [r]^t} \Lambda(\bk) \left(\prod_{\ell \in [t]} q_{\ell k_{\ell}}(x_{\ell}(i_{\ell})) - \prod_{\ell \in [t]} q_{\ell k_{\ell}}(x_{\ell}(i'_{\ell}))\right)\right| \\
&\leq \sum_{\ell \in [t]} \left|\sum_{\bk \in [r]^t} \Lambda(\bk) \prod_{\ell' < \ell} q_{\ell' k_{\ell'}}(x_{\ell'}) (q_{\ell k_{\ell}}(x_{\ell}(i_{\ell})) - q_{\ell k_{\ell}}(x_{\ell}(i'_{\ell}))) \prod_{\ell'' > \ell} q_{\ell'' k_{\ell''}}(x_{\ell''}(i_{\ell''})) \right| \\
&\leq \sum_{\ell \in [t]} B^{t-1}  \left| \sum_{h \in [r]} (q_{\ell h}(x_{\ell}(i_{\ell})) - q_{\ell h}(x_{\ell}(i'_{\ell})))\rho_{\ell}(h)\right| \\
&\leq \sum_{\ell \in [t]} B^{t-1} \sqrt{r \sum_{h \in [r]} (q_{\ell h}(x_{\ell}(i_{\ell})) - q_{\ell h}(x_{\ell}(i'_{\ell})))^2 \rho_{\ell}(h)^2} \\
&\leq B^{t-1} \sqrt{r} \sum_{\ell \in [t]} \max_{h \in [r]} |\rho_{\ell}(h)| \sqrt{\|Q_{\ell}^T (e_{i_{\ell}} - e_{i'_{\ell}})\|_2^2},
\end{align*}
where we used the assumption that $\max_{\ell, k} \sup_{x} |q_{\ell k}(x)| \leq B$.
By construction, $\hLam = \hU \hSig \hV^T$ and $\tQ_y = Q_y \hU$ where $\hU$ and $\hV$ are orthonormal such that
\begin{align*}
\|\hSig^{s+1} \tQ_y^T (e_a - e_{a'})\|_2^2 
&= (e_a - e_{a'})^T \tQ_y \hat{\Sigma}^{2(s+1)} \tQ_y^T (e_a - e_{a'}) \\
&= (e_a - e_{a'})^T Q_y (\hat{\Lambda} \hat{\Lambda}^T)^{(s+1)} Q_y (e_a - e_{a'}) \\
&\geq \hsig_{\min}^{2(s+1)} \|Q_y^T (e_a - e_{a'})\|_2^2.
\end{align*}
Let $\rho_{\max} := \max_{\ell \in [t] h \in [r]} |\rho_{\ell}(h)|$. It follows then that,
\begin{align}
|T(\bi') - T(\bi)| \leq B^{t-1} \sqrt{r} \rho_{\max} \hsig_{\min}^{-(s+1)} \sum_{\ell \in [t]} \|\hSig^{s+1} \tQ_{\ell}^T (e_{i_{\ell}} - e_{i'_{\ell}})\|_2.
\end{align}

By construction, if $K(\bi, \bi') > 0$, then for all $\ell \in [t]$, 
$\hat{d}_{\ell}(i_{\ell}, i'_{\ell}) \leq \eta$. Conditioned on the good event from Lemma \ref{lem:dist_conc}, it follows that for $\bi, \bi'$ such that $K(\bi, \bi') > 0$ and for all $\ell \in [t]$,
\[\|\hSig^{s+1} \tQ_{\ell}^T (e_{i_{\ell}} - e_{i'_{\ell}})\|_2^2 \leq \eta + \frac{5 r B \hsig_{\max}^{s+1} (1 + \hsig_{\max}^{s}) \log^{1/2}(trn^2) (1+o(1))}{(n\tp)^{1/2}}.\]
Let $C_{d} = 5 r B \hsig_{\max}^{s+1} (1 + \hsig_{\max}^{s})$.
Next we argue that the number of datapoints $\bi' \in \Omega_3$ such that $K(\bi, \bi') > 0$ is sufficiently large such that the noise term in the estimate is small. Conditioned on the good event in Lemmas \ref{lem:dist_conc} and \ref{lem:suff_many_nn} with
\[\eta' =  \eta - \frac{C_d \log^{1/2}(trn^2) (1+o(1))}{(n\tp)^{1/2}},\]
it follows that
\begin{align}
\sum_{\bi'} K(\bi, \bi') 
&\geq \sum_{\bi'} \prod_{\ell \in [t]} \Ind{\|\hSig^{s+1} \tQ_{\ell}^T (e_{i_{\ell}} - e_{i'_{\ell}})\|_2^2 \leq \eta - \frac{C_d \log^{1/2}(trn^2) (1+o(1))}{(n\tp)^{1/2}}} \nonumber \\
&= \prod_{\ell \in [t]} \left(1 + \sum_{i' \neq i_{\ell}} \Ind{\|\hSig^{s+1} \tQ_{\ell}^T (e_{i_{\ell}} - e_{i'_{\ell}})\|_2^2 \leq \eta - \frac{C_d \log^{1/2}(trn^2) (1+o(1))}{(n\tp)^{1/2}}}\right) \nonumber \\
&\geq \left(1 + \frac{(1-\delta) (n-1)}{\hsig_{\max}^s L} \sqrt{\eta - \frac{C_d \log^{1/2}(trn^2) (1+o(1))}{ (n\tp)^{1/2}}}\right)^t. \label{eq:tensor_neigh1}
\end{align}

We assumed that $\Omega_3$ is a freshly sampled dataset such that
$\sum_{\bi'} \Ind{\bi' \in \Omega_3} K(\bi, \bi')$
is distributed as a Binomial$\left(\sum_{\bi'} K(\bi, \bi'), p\right)$. By Chernoff's bound,
\begin{align}
\Prob\left(\sum_{\bi'} \Ind{\bi' \in \Omega_3} K(\bi, \bi') \leq (1-\gamma) p \sum_{\bi'} K(\bi, \bi')\right) \leq \exp\left(-\frac12 \gamma^2 p \sum_{\bi'} K(\bi, \bi')\right). \label{eq:tensor_neigh2}
\end{align}
Conditioned on $\Omega_3$, the noise terms $(T^{obs}_3(\bi') - T(\bi'))$ are independent, mean zero, and bounded in $[-1,1]$. By Hoeffding's bound,
\begin{align}
\Prob\left(\left|\frac{\sum_{\bi'} \Ind{\bi' \in \Omega_3} K(\bi, \bi') (T^{obs}_3(\bi') - T(\bi'))}{\sum_{\bi'} \Ind{\bi' \in \Omega_3} K(\bi, \bi')}\right| \geq \sqrt{\frac{2 \log(n^{t+1})}{\sum_{\bi'} \Ind{\bi' \in \Omega_3} K(\bi, \bi')}} \right) \leq \frac{2}{n^{t+1}}. \label{eq:obsnoise_bd}
\end{align}

As a result, conditioned on the good events in Lemmas \ref{lem:dist_conc} and \ref{lem:suff_many_nn}, with probability 
\[1- \frac{2}{n^{t+1}} - \exp\left(-\frac12 \gamma^2 p \left(1 + \frac{(1-\delta) (n-1)}{\hsig_{\max}^s L} \sqrt{\eta - \frac{C_d \log^{1/2}(trn^2) (1+o(1))}{(n\tp)^{1/2}}}\right)^t\right),\]
for constants $\delta, \gamma$, and for 
\[\eta \geq \frac{2 C_d \log^{1/2}(trn^2) (1+o(1))}{(n\tp)^{1/2}},\]
it holds that 
\begin{align*}
&|\hat{T}(\bi) - T(\bi)| \\
&= t B^{t-1} \sqrt{r} \rho_{\max} \hsig_{\min}^{-(s+1)} \sqrt{\eta + \frac{C_d \log^{1/2}(trn^2) (1+o(1))}{(n\tp)^{1/2}}} \\
&\quad + \sqrt{2 \log(n^{t+1})} \left((1-\gamma) p \left(1 + \frac{(1-\delta) (n-1)}{\hsig_{\max}^s L} \sqrt{\eta - \frac{C_d \log^{1/2}(trn^2) (1+o(1))}{(n\tp)^{1/2}}}\right)^t\right)^{-\frac12} \\
&= \Theta\left( t B^{t-1} \sqrt{r} \rho_{\max} \hsig_{\min}^{-(s+1)} \sqrt{\max\left(\eta, \frac{r B \hsig_{\max}^{2s+1} \log^{1/2}(n^2)}{n^{\kappa/2}}\right)}\right) + \Theta\left(\sqrt{\frac{\hsig_{\max}^{st} L^t \log(n^{t+1})}{pn^t \eta^{t/2}}}\right). 
\end{align*}
By choosing $\eta$ to balance the two terms, the bound is minimized for 
\begin{align}
\eta = \Theta\left(\max\left(\frac{\log^{1/2}(n^2)}{(n\tp)^{1/2}}, \frac{\log^{2/(t+2)}(n^{t+1})}{n^{2(\kappa+1)/(t+2)}}\right)\right).\label{eq:eta_def}
\end{align}
Recall that $n\tp = \Theta(n^{\min(\kappa,1)})$. For this choice of $\eta$, it follows by \eqref{eq:tensor_neigh1} that with high probability 
\begin{align*}
p \sum_{\bi'} K(\bi, \bi') &= \Theta\left(p (n \sqrt{\eta'})^t\right) 
= \Omega\left(\frac{\log(n^{t+1})}{\eta}\right) 
= \Omega(n^{\min(\kappa/2, 1/2, 2(\kappa+1)/(t+2))}).
\end{align*}
We can plug in this bound to simplify \eqref{eq:tensor_neigh2}. Given this choice of $\eta$, we can also simplify the probability of error in Lemma \ref{lem:suff_many_nn} by the fact that
\[n \sqrt{\eta} = \Theta\left(\max\left(n^{1-\min(\kappa,1)/4} \log^{1/4}(n^2), n^{1-(\kappa+1)/(t+2)} \log^{1/(t+2)}(n^{t+1})\right)\right) = \Omega(n^{1/2}).\]

As we would like to show the entrywise error bound over all $n^t$ entries, we take the intersection of all the good events and apply union bound. We use Lemma \ref{lem:dist_conc} for each of the $\ell \in [t]$ modes. We use \ref{lem:suff_many_nn} for each of the $\ell \in [t]$ modes and $i \in [n]$ coordinates. We union bound over all $\bi \in [n]^t$ entries for the bounds in \eqref{eq:tensor_neigh2} and \eqref{eq:obsnoise_bd}. It follows that for constant $\delta, \gamma$, and $\eta$ chosen according to \eqref{eq:eta_def}, with probability at least 
\begin{align*}
1& - \frac{6t(1+o(1))}{tn} - t n \exp\left(-\frac{\delta^2 (n-1)\sqrt{\eta'}}{2 \hsig_{\max}^s L}\right) - n^t \exp\left(-\frac12 \gamma^2 p \sum_{\bi'} K(\bi, \bi')\right) - n^t \left(\frac{2}{n^{t+1}}\right) \\
&\qquad= 1 - \frac{8(1+o(1))}{n} - t n \exp\left(-\Omega(n^{1/2})\right) - n^t \exp\left(-n^{\min(\kappa/2, 1/2, 2(\kappa+1)/(t+2))}\right) \\
&\qquad= 1 - \frac{8(1+o(1))}{n},
\end{align*}
the estimate output by our algorithm satisfies 
\[\max_{\bi \in [n]^t} |\hat{T}(\bi) - T(\bi)| = O\left(\max\left(\frac{\log^{1/4}(n^2)}{n^{\min(\kappa,1)/4}}, \frac{\log^{1/(t+2)}(n^{t+1})}{n^{(\kappa+1)/(t+2)}}\right)\right).\]
It follows then that 
\[\E{\frac{1}{n^t} \sum_{\bi \in [n]^t} (\hat{T}(\bi) - T(\bi))^2} = O\left(\max\left(\frac{\log^{1/2}(n^2)}{n^{\min(\kappa,1)/2}}, \frac{\log^{2/(t+2)}(n^{t+1})}{n^{2(\kappa+1)/(t+2)}}\right)\right).\]

\end{new-proof}

\subsection{Sufficiently Many Nearest Neighbors} \label{sec:suff_many_nn}

In this section, we prove Lemma \ref{lem:suff_many_nn}, which states that there are sufficiently many nearest neighbors. 

\begin{new-proof}{Lemma \ref{lem:suff_many_nn}}
By construction, 
\begin{align*}
	\|\hSig^{s+1} \tQ_y^T (e_a - e_{a'})\|_2^2
	&\leq \hsig_{\max}^{2s}  \|\hSig \tQ_y^T (e_a - e_{a'})\|_2^2\\
	&= \hsig_{\max}^{2s} \int_{0^1} (\tilde{f}(x_y(a), x') - \tilde{f}(x_y(a'), x'))^2 d x' \\
	&\leq  \hsig_{\max}^{2s} L^2 (x_y(a) - x_y(a'))^2
\end{align*}
where the last inequality follows from Lipschitzness of $\tilde{f}$ as shown in \eqref{eq:tf_Lipschitz}.
As a result, for any $\eta' > 0$, if $|x_{\ell}(i) - x_{\ell}(i')| \leq \frac{\sqrt{\eta'}}{\hsig_{\max}^{s} L}$, then $	\|\hSig^{s+1} \tQ_y^T (e_a - e_{a'})\|_2^2 \leq \eta'$. 
By the model assumption that for $i \in [n]$, the latent variables $x_{\ell}(i)$ are sampled i.i.d., it follows that for any $i \in [n]$,
\[\sum_{i' \neq i \in [n]} \Ind{	\|\hSig^{s+1} \tQ_y^T (e_a - e_{a'})\|_2^2 \leq \eta'}\]
stochastically dominates a Binomial$\left(n-1, \frac{\sqrt{\eta'}}{\hsig_{\max}^{s} L}\right)$ distributed random variable. Therefore the lemma statement follows by Chernoff's bound.
\end{new-proof}

\subsection{Concentration of Distance Estimates} \label{sec:dist_conc}

This section focuses on proving the key Lemma \ref{lem:dist_conc}, which shows that the estimated distances concentrate well. We will assume $p = n^{-(t-1) + \kappa}$ for $\kappa > 0$; it follows then that $s = \lceil \frac{\ln(n)}{\ln(pn^{t-1})} \rceil = \lceil \frac{1}{\kappa} \rceil$. When $\kappa < 1$, then $\tp = \Theta(n^{\kappa - 1})$ such that $\tp = o(1)$. When $\kappa \geq 1$, then $\tp = \Theta(1)$ and the constructed matrix $M^{obs}_{yz}$ is dense.

Let us denote event 
\[\cA^1_{a,s}(\delta,C) = \cup_{\ell=1}^{s-1} \{|\cS_{a, \ell}| \in [((1-\delta) n\tp)^{\ell}(1-o(1)), ((1+\delta)n\tp)^{\ell}]\} \cup \{|\cS_{a,s}| \geq C n\} \cup \{|\cS_{a,s+1}| \geq C n\}.\]

\begin{lemma}\label{lem:neighborhood_growth}
Assume $p = n^{-(t-1) + \kappa}$ for $\kappa > 0$. For $s = \lceil \frac{1}{\kappa} \rceil$ and any $\delta \in (0,1)$ and $a \in [n]$, there exists some constant $C$ such that 
\begin{align*}
\Prob\left(\neg \cA^1_{a,s}(C)\right) \leq 2\exp\left(-\frac{\delta^2 (n \tp) (1 - o(1))}{3}\right).
\end{align*}
\end{lemma}
The proof is deferred to Section \ref{sec:nhbrhd}.

Recall that $\hsig_k$ denotes the $k$-th singular value of the matrix $\hat{\Lambda}$ defined in \eqref{eq:hat_Lambda}, and $\hat{\Sigma}$ is the diagonal matrix with $\hsig_k$ on the $k$-th diagonal entry. 

Let us denote event
\begin{align*}
\cA^2_{a,j,k}(\delta) = \begin{cases}
\left\{|e_k^T \tQ_y \tN_{a,j} - e_k^T \hSig^j \tQ_y e_a| \leq \frac{\hsig_k^{j-1} \log^{1/2}(trn^2)}{((1-\delta)n\tp)^{1/2}}\right\} &\text{ if } j \text{ is even}, \\
\left\{|e_k^T \tQ_z \tN_{a,j} - e_k^T \hSig^j \tQ_y e_a| \leq \frac{\hsig_k^{j-1} \log^{1/2}(trn^2)}{((1-\delta)n\tp)^{1/2}}\right\} &\text{ if } j \text{ is odd}.
\end{cases}
\end{align*}

\begin{lemma}\label{lem:martingale}
Assume $p = n^{-(t-1) + \kappa}$ for $\kappa > 0$. For $s = \lceil \frac{1}{\kappa} \rceil$, constants $\delta \in (0,1)$ and $C$, and any $a \in [n]$, $k \in [r]$, $j \in [s+1]$,
\begin{align*}
\Prob\left(\neg \cA^2_{a,j,k}(\delta) ~|~ \cA^1_{a,s}(\delta,C), \left\{\{x_{\ell}(i)\}_{i \in [n]}\right\}_{\ell \in [t] \setminus \{y,z\}}\right) \leq \frac{2(1+o(1))}{trn^2},
\end{align*}
where $\cT = [t] \setminus \{y,z\}$.
\end{lemma}
The proof is deferred to Section \ref{sec:martingale}.

If $s$ is even, let us denote event 
\begin{align*}
\cA^3_{a,a'} = \Bigg\{&\left|D(a,a') - \tN_{a,s}^T \tQ_y^T \hSig \tQ_z \tN_{a',s+1}\right| \\
&\qquad\leq  \max\left(\frac{2 \log^{1/2}(t n^3)}{(p n^{t-2} |\cS_{a,s}| |\cS_{a',s+1}|)^{1/2}}, \frac{8 \log(tn^3)}{3 |\cS_{a,s}| |\cS_{a',s+1}|}\right) (1+o(1)) \Bigg\}.
\end{align*}
and if $s$ is odd, let us denote event 
\begin{align*}
\cA^3_{a,a'} = \Bigg\{&\left|D(a,a') - \tN_{a,s}^T \tQ_z^T \hSig \tQ_y \tN_{a',s+1}\right| \\
&\qquad\leq  \max\left(\frac{2 \log^{1/2}(t n^3)}{(p n^{t-2} |\cS_{a,s}| |\cS_{a',s+1}|)^{1/2}}, \frac{8 \log(tn^3)}{3 |\cS_{a,s}| |\cS_{a',s+1}|}\right) (1+o(1)) \Bigg\}.
\end{align*}

\begin{lemma}\label{lem:inner_prod}
Assume $p = n^{-(t-1) + \kappa}$ for $\kappa > 0$. For $s = \lceil \frac{1}{\kappa} \rceil$, constants $\delta \in (0,1)$ and $C$, and any $a,a' \in [n]$, it holds that 
\begin{align*}
&\Prob\left(\neg \cA^3_{a,a'}(\phi) ~|~ \cA^1_{a,s}(\delta,C), \cap_{k=1}^r \cA^2_{a,s,k}(\delta), \cA^1_{a',s}(\delta,C), \cap_{k=1}^r \cA^2_{a',s+1,k}(\delta), \{\{x_{\ell}(i)\}_{i \in [n]}\}_{\ell \in [t] \setminus \{y,z\}}\right) \\
&\qquad\qquad\leq \frac{2}{t n^3} + \frac{1}{n^6}.
\end{align*}
\end{lemma}
The proof is deferred to Section \ref{sec:innerprod}.

\begin{new-proof}{Lemma \ref{lem:dist_conc}}
Conditioned on events $\cA^1_{a,s}(\delta,C), \cA^1_{a',s}(\delta,C), \cap_{k=1}^r \cA^2_{a,s,k}(\delta),\cap_{k=1}^r\cA^2_{a',s+1,k}(\delta)$, if $s$ is even, 
\begin{align*}
&|\tN_{a,s}^T \tQ_y^T \hSig \tQ_z \tN_{a',s+1} - e_a^T \tQ_y^T \hSig^{2s +2} \tQ_y e_{a'}| \\
&= \left|\sum_k \hsig_k (e_k^T \tQ_y \tN_{a,s}) (e_k^T \tQ_z \tN_{a',s+1} - e_k^T \hSig^{s+1} \tQ_y e_{a'})
+ \sum_k \hsig_k (e_k^T \tQ_y \tN_{a,s} - e_k^T \hSig^s \tQ_y e_a) (e_k^T \hSig^{s+1} \tQ_y e_{a'})\right| \\
&\leq B \sum_k \frac{\hsig_k^{s+1} \log^{1/2}(trn^2)}{((1-\delta)n\tp)^{1/2}}
+ B \sum_k \frac{\hsig_k^{2s+1} \log^{1/2}(trn^2)}{((1-\delta)n\tp)^{1/2}} \\
&\leq \frac{r B \hsig_{\max}^{s+1} (1 + \hsig_{\max}^{s}) \log^{1/2}(trn^2)}{((1-\delta)n\tp)^{1/2}}.
\end{align*}
When $s$ is odd, a similar argument shows that 
\begin{align*}
|\tN_{a,s}^T \tQ_z^T \hSig \tQ_y \tN_{a',s+1} - e_a^T \tQ_y^T \hSig^{2s +2} \tQ_y e_{a'}| 
&\leq \frac{r B \hsig_{\max}^{s+1} (1 + \hsig_{\max}^{s}) \log^{1/2}(trn^2)}{((1-\delta)n\tp)^{1/2}}.
\end{align*}

To put it all together, conditioned on events $\cA^1_{a,s}(\delta,C), \cA^1_{a',s}(\delta,C), \cap_{k=1}^r (\cA^2_{a,s,k}(\delta) \cap \cA^2_{a,s+1,k}(\delta))$, $\cap_{k=1}^r (\cA^2_{a',s,k}(\delta) \cap \cA^2_{a',s+1,k}(\delta)), \cA^3_{a,a'}(\phi), \cA^3_{a,a}(\phi), \cA^3_{a',a}(\phi), \cA^3_{a',a'}(\phi)$, 
for constant $\delta$,
\begin{align*}
|\hat{d}_y(a,a') - \|\hSig^{s+1} \tQ_y^T (e_a - e_{a'})\|_2^2| 
&\leq \max\left(\frac{8 \log^{1/2}(t n^3)}{(p n^{t-2} |\cS_{a,s}| |\cS_{a',s+1}|)^{1/2}}, \frac{32 \log(tn^3)}{3 |\cS_{a,s}| |\cS_{a',s+1}|}\right) (1+o(1)) \\
&\qquad + \frac{4 r B \hsig_{\max}^{s+1} (1 + \hsig_{\max}^{s}) \log^{1/2}(trn^2)}{((1-\delta)n\tp)^{1/2}}
\end{align*}
The first term scales as $\Theta\left(\max(\frac{\log^{1/2}(t n^3)}{n^{(1+\kappa)/2}},\frac{\log(t n^3)}{n^2})\right)$ and the second term scales as $\Theta\left(\frac{\log^{1/2}(trn^2)}{\min(n^{\kappa/2},n^{1/2})}\right)$
The first term is always dominated by the second term. We plug in a choice of $\delta = \frac{9}{25}$ to get the final bound.

To guarantee this bound on the distance estimates for all pairs $a,a' \in [n]^2$, we take the intersection of the good events over pairs $a,a' \in [n]^2$. The max bound on the distance estimates hold when events $\cap_{a \in [n]} \cA^1_{a,s}(\delta,C), \cap_{a \in [n]} \cap_{k=1}^r (\cA^2_{a,s,k}(\delta) \cap \cA^2_{a,s+1,k}(\delta)), \cap_{a, a' \in [n]^2} \cA^3_{a,a'}(L)$ 
hold. By Lemmas \ref{lem:neighborhood_growth}, \ref{lem:martingale}, \ref{lem:inner_prod} and union bound, for constant $\delta$, these good events hold with probability at least 
\begin{align*}
1 - 2 n \exp\left(-\frac{\delta^2 n^{\kappa} (1 - o(1))}{3}\right)
- \frac{4rn(1+o(1))}{trn^2} - \frac{2 n^2}{t n^3} - \frac{1}{n^6} = 1 - \frac{6(1+o(1))}{t n}.
\end{align*}

\end{new-proof}

\subsection{Rate of Local Neighborhood Growth} \label{sec:nhbrhd}

\begin{new-proof}{Lemma \ref{lem:neighborhood_growth}}
We first handle the setting that $p = n^{-(t-1) + \kappa}$ for $\kappa \geq 1$. By definition,
\[\tp = (1- (1-p)^{n^{t-2}}) = (1 - (1-\frac{n^{\kappa-1}}{n^{t-2}})^{n^{t-2}}).\]
Recall that $e^{-x} = \lim_{n\to\infty} (1 + \frac{-x}{n})^n$. Therefore for $\kappa = 1$, $\tp \to 1-e^{-1} = \Theta(1)$. For $\kappa > 1$, then $\tp \to 1 = \Theta(1)$. As a result, in the setting where $\kappa \geq 1$, the density of observations $\tp$ in our constructed matrix is constant. For $s=1$, $|\cS_{a,s}| \sim \text{Binomial}(n, \tp)$. By Chernoff's bound, it holds that for $C = (1-\delta) \tp = \Theta(1)$,
\[\Prob\left(|\cS_{a,s}| \geq Cn\right) \geq 1 - \exp\left(-\frac{\delta^2 (n \tp)}{3}\right).\]
Conditioned on $|\cS_{a,s}| \geq Cn$, $|\cS_{a,s+1}| \sim \text{Binomial}(n-1, 1 - (1-\tp)^{|\cS_{a,s}|})$. As $1-\tp < 1$ and $|\cS_{a,s}| = \omega(1)$, then $1 - (1-\tp )^{|\cS_{a,s}|} \to 1$. By Chernoff's bound, it holds that for $C = (1-\delta) \tp = \Theta(1)$,
\[\Prob\left(\left. |\cS_{a,s+1}| \geq C n ~\right|~ |\cS_{a,s}| \geq C n\right) \geq 1 - \exp\left(-\frac{(1-C)^2 n}{3}\right).\]
For $\delta \in (0,1)$, we can verify that $(1-C)^2 \geq \delta$.

We next address the ultra-sparse setting where $p = n^{-(t-1) + \kappa}$ for $\kappa \in (0,1)$, such that $\tp = 1 - (1-p)^{n^{t-2}} = \Theta(n^{\kappa - 1}) = o(1)$. Recall that our graph is bipartite between vertex sets $\cV_1 = [n]$ and $\cV_2 = [n]$. Without loss of generality, assume that $a \in \cV_1$. Let $\cF_{a,h}$ denote the sigma-algebra containing information about the latent parameters, edges and the values associated with vertices in the bipartite graph up to distance $h$ from $a$, i.e. the depth $h$ radius neighborhood of $a$.

Let $p_{a,\ell} = 1-(1-\tp)^{|\cS_{a,\ell-1}|}$, and let
\begin{align*}
n_{a, \ell} = \begin{cases}
|\cV_2 \setminus \cup_{i=0}^{\lfloor\ell/2\rfloor - 1} \cS_{a,2i+1}| &\text{ if } \ell \text{ is odd} \\
|\cV_1 \setminus \cup_{i=0}^{\ell/2 - 1} \cS_{a,2i}| &\text{ if } \ell \text{ is even}.
\end{cases}
\end{align*}

For depth $\ell$, conditioned on $\cF_{\ell-1}$, $|\cS_{a, \ell}|$ is distributed according to a Binomial with parameters $(n_{a,\ell},p_{a,\e})$. It follows by Chernoff's bound that
\[\Prob(|\cS_{a, \ell}| \notin (1 \pm \delta) n_{a,\ell} p_{a,\ell} ~|~ \cF_{a,\ell-1}) \leq 2\exp\left(-\frac{\delta^2 n_{a,\ell} p_{a,\ell}}{3}\right).\]

Let us define the following event $\tilde{\cA}_{a,h}$,
\[\tilde{\cA}_{a,h} = \cap_{\ell=1}^{h} \{|\cS_{a, \ell}| \notin (1 \pm \delta) n_{a,\ell} p_{a,\ell}\}.\]

Next we argue that for $s = \lceil \frac{1}{\kappa} \rceil$, event $\tilde{\cA}_{a,s}$ implies that for all $\ell \in [s-1]$, 
\[\{|\cS_{a, \ell}| \in [((1-\delta) n\tp)^{\ell}(1-o(1)), ((1+\delta)n\tp)^{\ell}]\},\]
$|\cS_{a,s}| = \Theta(n)$, and $|\cS_{a,s+1}| = \Theta(n)$.

We first prove the upper bounds on $|\cS_{a,\ell}|$. Naively, $|\cS_{a,\ell}| \leq n$. Conditioned on $\tilde{\cA}_{a,s}$,
\begin{align*}
|\cS_{a,\ell}| &\leq (1 + \delta) n_{a,\ell} p_{a,\ell} \\
&\leq (1 + \delta) n \left(1-(1-\tp)^{|\cS_{a,\ell-1}|}\right) \\
&\leq (1 + \delta) n \tp |\cS_{a,\ell-1}|.
\end{align*}
By inductively repeating this argument and using the fact that $|\cS_{a,0}| = 1$, it follow that 
\[|\cS_{a,\ell}| \leq ((1 + \delta) n \tp)^{\ell}.\]
For $\ell < \frac{1}{\kappa}$, $(n \tp)^{\ell} = o(n)$.

Next we prove the lower bounds on $|\cS_{a,\ell}|$ for $\ell \in [s+1]$. Let us assume without loss of generality that $\ell$ is even (same argument holds for $\ell$ odd). Conditioned on $\tilde{\cA}_{a,s}$,
\begin{align*}
n_{a,\ell} &\geq n - \sum_{i=0}^{\ell/2 - 1} |\cS_{a,2i}|  \\
&\geq n - \sum_{i=0}^{\ell/2 - 1} ((1 + \delta) n \tp)^{2i} \\
&= n(1-o(1))
\end{align*}
The last step follows from the fact that $2i \leq \ell - 2 \leq \lceil \frac{1}{\kappa} \rceil-1 < \frac{1}{\kappa}$, such that $((1 + \delta) n \tp)^{2i} = o(1)$. Furthermore our choice of $s$ guarantees it to be a constant (as $\kappa$ is constant). Next we want to lower bound $p_{a,\ell}$. Conditioned on $\tilde{\cA}_{a,s}$,
\begin{align*}
p_{a,\ell} &= 1-(1-\tp)^{|\cS_{a,\ell-1}|} \geq \tp |\cS_{a,\ell-1}| (1- \tp |\cS_{a,\ell-1}|)
\end{align*}
This lower bound is only useful when $\tp |\cS_{a,\ell-1}| = o(1)$, otherwise the bound could be negative. By the upper bound on $|\cS_{a,\ell-1}|$, for $\ell < s$, it holds that $\tp |\cS_{a,\ell-1}| \leq \frac{1}{n} (\tp n)^{\ell} = o(1)$. Therefore, for $\ell < s$,
\begin{align*}
|\cS_{a,\ell}| &\geq (1 - \delta) n_{a,\ell} p_{a,\ell} \\
&\geq (1 - \delta) n(1-o(1)) \tp |\cS_{a,\ell-1}| (1- \tp |\cS_{a,\ell-1}|) \\
&\geq (1 - \delta) n(1-o(1)) \tp |\cS_{a,\ell-1}| (1 - o(1)).
\end{align*}
By inductively repeating this argument and using the fact that $|\cS_{a,0}| = 1$, it follow that for $\ell < s$,
\[|\cS_{a,\ell}| \geq ((1 + \delta) n \tp)^{\ell} (1-o(1)).\]

Next we prove that $|\cS_{a,\ell}| = \Omega(n)$ for $\ell \in \{s,s+1\}$. Recall that $e^{-x} = \lim_{n\to\infty} (1 + \frac{-x}{n})^n$. For $\ell \geq s \geq \frac{1}{\kappa}$, using the fact that $\tp = \Theta(n^{\kappa-1})$,
\begin{align*}
\tp |\cS_{a,\ell-1}| &\geq \tp ((1 + \delta) n \tp)^{\frac{1 - \kappa}{\kappa}} = \Theta(n^{\kappa-1} ((1 + \delta) n^{\kappa})^{\frac{1 - \kappa}{\kappa}}) = \Theta(1), \text{ for some constant } > 0.
\end{align*}
As a result, 
\[\lim_{n\to\infty} \left(1-\frac{\tp |\cS_{a,\ell-1}|}{|\cS_{a,\ell-1}|}\right)^{|\cS_{a,\ell-1}|} \leq e^{-\tp |\cS_{a,\ell-1}|} < 1,\]
which implies that for $\ell \in \{s,s+1\}$, $p_{a,\ell} \geq C$ for some constant $C > 0$.
Therefore, for $\ell < s$,
\begin{align*}
|\cS_{a,\ell}| &\geq (1 - \delta) n_{a,\ell} p_{a,\ell} \\
&\geq (1 - \delta) n(1-o(1)) C \\
&= \Omega(n).
\end{align*}

To complete the proof, we use the lower bounds on $|\cS_{a,\ell}|$ to reduce the probability bounds. 
\begin{align*}
\Prob(\neg \tilde{\cA}_{a,s+1})
&= \Prob\left(\cup_{\ell=1}^{s+1} \{|\cS_{a, \ell}| \notin (1 \pm \delta) n_{a,\ell} p_{a,\ell}\}\right) \\
&\leq \sum_{\ell=1}^{s+1} \Prob\left(|\cS_{a, \ell}| \notin (1 \pm \delta) n_{a,\ell} p_{a,\ell} ~|~ \tilde{\cA}_{a,\ell-1}\right) \\
&\leq \sum_{\ell=1}^{s+1} 2\exp\left(-\frac{\delta^2 n_{a,\ell} p_{a,\ell}}{3}\right) \\
&\leq \sum_{\ell=1}^{s-1} 2\exp\left(-\frac{\delta^2 (n \tp)^{\ell} (1 - o(1))}{3}\right) + 4\exp\left(-\frac{\delta^2 n C}{3}\right).
\end{align*}
Note that the first term significantly dominates the remaining terms, as $n \tp$ is asymptotically smaller than $n$ and $(n \tp)^{\ell}$ for any $\ell > 1$. Therefore, the remaining terms get absorbed by the $o(1)$ in the exponent, leading to
\begin{align*}
\Prob(\neg \tilde{\cA}_{a,s+1}) \leq 2\exp\left(-\frac{\delta^2 (n \tp) (1 - o(1))}{3}\right).
\end{align*}

\end{new-proof}
\subsection{Martingale Concentration of Neighborhood Statistics} \label{sec:martingale}
\begin{new-proof}{Lemma \ref{lem:martingale}}
Assume without loss of generality that $a \in \cV_1$. Let us define
\begin{align*}
Y_{a,h} &= \begin{cases}
e_k^T \hSig^{s+1-h} \tQ_y^T \tN_{a,h} &\text{ if } h \text{ is even} \\
e_k^T \hSig^{s+1-h} \tQ_z^T \tN_{a,h} &\text{ if } h \text{ is odd}
\end{cases} \\
D_{a,h} &= Y_{a,h} - Y_{a,h-1}
\end{align*}
so that $\sum_{h=1}^{j} D_{a,h} = Y_{a,j} - e_k^T \hSig^{s+1} \tQ_y^T e_a$.
Conditioned on $\cS_{a, \ell}$ for all $\ell \in [s+1]$ and conditioned on all $\left\{\{x_{\ell}(i)\}_{i \in [n]}\right\}_{\ell \in [t] \setminus \{y,z\}}$, let $\cF_{a,h}$ denote the sigma-algebra containing information about the latent parameters, edges and the values associated with vertices in the bipartite graph up to distance $h$ from $a$, i.e. the depth $h$ radius neighborhood of $a$. This includes $x_{y}(i)$ and $x_z(i)$ for all $i \in \cup_{\ell = 0}^h \cS_{a,h}$, as well as $M^{obs}_{yz}(i,j)$ for any edge $i,j$ such that $i$ or $j$ is at distance at most $h-1$ from vertex $a$. Conditioned on $\cF_{a,h}$, the BFS tree rooted at vertex $a$ up to depth $h$ is measurable, as are quantities $\tN_{a,\ell}$ for any $\ell \leq h$.

We will show that conditioned on $\cS_{a, \ell}$ for all $\ell \in [s+1]$ and conditioned on all $\left\{\{x_{\ell}(i)\}_{i \in [n]}\right\}_{\ell \in [t] \setminus \{y,z\}}$, $\{(D_{a,h}, \cF_h)\}$ is a martingale difference sequence with controlled conditional variances such that martingale concentration holds.


Without loss of generality, let's assume that $h$ is even (the below argument will also follow for odd $h$)
\begin{align*}
D_{a,h} &= e_k^T \hSig^{s+1-h} \tQ_y^T \tN_{a,h} - e_k^T \hSig^{s-h} \tQ_z^T \tN_{a,h-1} \\
&= \hsig_k^{s+1-h} (e_k^T \tQ_y^T \tN_{a,h} - e_k^T \hSig \tQ_z^T \tN_{a,h-1}) \\
&= \frac{\hsig_k^{s+1-h}}{|\cS_{a,h}|} \sum_{i \in \cS_{a,h}} (N_{a,h}(i) \tilde{q}_{yk}(x_y(i)) - e_k^T \hSig \tQ_z^T \tN_{a,h-1}).
\end{align*}

$D_{a,h}$ can be written as a sum of independent terms $X_i$ for $i \in \cS_{a,h}$,
\begin{align*}
X_i &=  \frac{\hsig_k^{s+1-h}}{|\cS_{a,h}|} \left(N_{a,h}(i) \tilde{q}_{yk}(x_y(i)) - e_k^T \hSig \tQ_z^T \tN_{a,h-1}\right) \\
&=  \frac{\hsig_k^{s+1-h}}{|\cS_{a,h}|} \left(\sum_{b \in \cS_{a,h-1}} N_{a,h-1}(b) \Ind{b = \pi(i)} M^{obs}_{yz}(i,b) \tilde{q}_{yk}(x_y(i)) - e_k^T \hSig \tQ_z^T \tN_{a,h-1}\right),
\end{align*}
where $\pi(i)$ denote the parent of $i$ in the BFS tree.
Conditioned on $\cF_{u,h-1}$, for any $i \in \cS_{a,h}$, any coordinate $b \in \cS_{a,h-1}$ is equally likely to be the parent of $i$ in the BFS tree due to the symmetry/uniformity of the sampling process. As a result, conditioned on $\cF_{u,h-1}$ and $i \in \cS_{a,h}$,
\begin{align*}
\E{X_i} &=  \frac{\hsig_k^{s+1-h}}{|\cS_{a,h}| |\cS_{a,h-1}|} \sum_{b \in \cS_{a,h-1}} N_{a,h-1}(b) \left(\E{M^{obs}_{yz}(i,b) \tilde{q}_{yk}(x_y(i)) | (i,b) \in \tilde{\Omega}_{yz}} - \tilde{q}_{zk}(x_z(b)) \hsig_k\right).
\end{align*}

First we verify that 
\begin{align*}
	&\E{\left. \frac{  \Ind{\bh \in \Omega_1}}{|\Omega_1 \cap \cI_{yz}(i,b)|} ~\right|~ (i,b) \in \tilde{\Omega}_{yz}} \\
	&\qquad= \sum_{g=0}^{n^{t-2}-1} \frac{1}{1+g} \Prob\left( \left.\bh \in \Omega_1, |\Omega_1 \cap \cI_{yz}(i,b)| = g+1 ~\right|~ |\Omega_1 \cap \cI_{yz}(i,b)| \geq 1\right) \\
	&\qquad= \sum_{y=0}^{n^{t-2}-1} \frac{1}{1+y} \frac{p \binom{n^{t-2}-1}{g} p^g (1-p)^{n^{t-2}-1-g}}{1 - (1-p)^{n^{t-2}}} \\
	&\qquad= \frac{p}{1 - (1-p)^{n^{t-2}}} \sum_{g=0}^{n^{t-2}-1} \frac{1}{p n^{t-2}} \binom{n^{t-2}}{g+1} p^{g+1} (1-p)^{n^{t-2}-(g+1)} \\
	&\qquad= \frac{p}{1 - (1-p)^{n^{t-2}}} \frac{1 - (1-p)^{n^{t-2}}}{p n^{t-2}} \\
	&\qquad= \frac{1}{n^{t-2}}.
\end{align*}
Conditioned on $\left\{\{x_{\ell}(i)\}_{i \in [n]}\right\}_{\ell \in[t] \setminus \{y,z\}}$, i.e. all latent variables in modes $3, 4, \dots t$, by \eqref{eq:Ex_M_obs}, 
\begin{align*}
\E{M^{obs}_{yz}(i,b) \tilde{q}_{yk}(x_y(i)) \mid (i,b) \in \tilde{\Omega}_{yz}} 
&= \sum_{h \in [r]} \hsig_h \tilde{q}_{yh}(x_y(i)) \tilde{q}_{yk}(x_y(i)) \tilde{q}_{zh}(x_z(b)) \\
&= \hsig_{k} \tilde{q}_{zk}(x_z(b)),
\end{align*}
implying that $\E{X_i} = 0$. Furthermore, as $\|N_{a,h}\|_{\infty} \leq 1$, 
\[|X_i| \leq \frac{\tB \hsig_k^{s+1-h} (1 + |\hsig_k|)}{|\cS_{a,h}|}.\]

Therefore $\{(D_{a,h}, \cF_h)\}$ is a martingale difference sequence with uniformly bounded differences. Next we want to establish concentration. Using the model assumptions that $|T^{obs}(\bi)| \leq 1$ such that $|M^{obs}_{yz}(i,b)| \leq 1$ and $\|N_{a,h}\|_{\infty} \leq 1$, it follows that 
\begin{align*}
\Var[X_i ~|~ i \in \cS_{a,h}]
&=  \frac{\hsig_k^{2(s+1-h)}}{|\cS_{a,h}|^2}  \Var\left[\sum_{b \in \cS_{a,h-1}} N_{a,h-1}(b) \Ind{b = \pi(i)} M^{obs}_{yz}(i,b) \tilde{q}_{yk}(x_y(i))\right] \\
&\leq \frac{\hsig_k^{2(s+1-h)}}{|\cS_{a,h}|^2} \E{\sum_{b \in \cS_{a,h-1}} \Ind{b = \pi(i)} N^2_{a,h-1}(b) (M^{obs}_{yz}(i,b))^2 (\tilde{q}_{yk}(x_y(i)))^2} \\
&\leq \frac{\hsig_k^{2(s+1-h)}}{|\cS_{a,h}|^2} \E{\sum_{b \in \cS_{a,h-1}} \frac{1}{|\cS_{a,h-1}|}} \\
&= \frac{\hsig_k^{2(s+1-h)}}{|\cS_{a,h}|^2}.
\end{align*}
Conditioned on $\{\cS_{a, \ell}\}_{\ell \in [s+1]}$ and $\left\{\{x_{\ell}(i)\}_{i \in [n]}\right\}_{\ell \in[t] \setminus \{y,z\}}$,
it follows that $D_{u,h}$ conditioned on $\cF_{u,h-1}$ is sub-exponential with parameters
\[\left(\frac{\hsig_k^{(s+1-h)}}{\sqrt{|\cS_{a,h}|}}, \frac{\tB \hsig_k^{s+1-h} (1 + |\hsig_k|)}{|\cS_{a,h}|} \right).\]
Conditioned on the event $\cA^1_{a,s}(\delta,C)$, the quantity $|\cS_{a,h}|$ can be lower bounded so that the sub-exponential parameters are bounded above by
\[\left(\frac{\hsig_k^{(s+1-h)}}{\sqrt{((1-\delta) n\tp)^{h}(1-o(1))}}, \frac{\tB \hsig_k^{s+1-h} (1 + |\hsig_k|)}{((1-\delta) n\tp)^{h}(1-o(1))} \right)\]
for $h \in [s-1]$, and 
\[\left(\frac{\hsig_k^{(s+1-h)}}{\sqrt{C n}}, \frac{\tB \hsig_k^{s+1-h} (1 + |\hsig_k|)}{C n} \right)\]
for $h \in \{s,s+1\}$.

By the Bernstein style bound for martingale concentration, it holds that $\sum_{h=1}^j D_{a,j}$ is sub-exponential with parameters $\nu_*$ and $\alpha_*$ for 
\begin{align*}
\nu_* &= \sqrt{\sum_{h=1}^{\min(s-1,j)} \frac{\hsig_k^{2(s+1-h)}}{((1-\delta) n\tp)^{h}(1-o(1))} + \frac{\hsig_k^{2} \Ind{s \leq j}}{C n} + \frac{\Ind{s+1 \leq j}}{C n}}\\
&= \frac{\hsig_k^{s} (1+ o(1))}{((1-\delta) n\tp)^{1/2}} \\
\alpha_* &= \max\left(\max_{h\in[s-1]} \frac{\tB \hsig_k^{s+1-h} (1 + |\hsig_k|)}{((1-\delta) n\tp)^{h}(1-o(1))}, \frac{\tB \hsig_k (1 + |\hsig_k|)}{Cn},\frac{\tB  (1 + |\hsig_k|)}{Cn}\right) \\
&= \frac{\tB \hsig_k^{s} (1 + |\hsig_k|) (1+o(1))}{(1-\delta) n\tp}
\end{align*}
where we use the fact that for sufficiently large $n$, $((1-\delta) n\tp)^{-1}$ asymptotically dominates $((1-\delta) n\tp)^{-h}$ for any $h > 1$. For the setting where $\kappa \geq 1$ and $s = 1$, we choose constant $C = (1-\delta)\tp$ such that $\nu_*$ and $\alpha_*$ also scale as the expressions above. It follows by Bernstein's inequality that 
for $0 < z < \frac{\hsig_k^{s} (1 - o(1))}{\tB (1 + |\hsig_k|)}$,
\begin{align*}
\Prob\left(|Y_{a,j} - e_k^T \hSig^{s+1} \tQ_y^T e_a| \geq \theta ~|~ \cA^1_{a,s}(\delta,C), \left\{\{x_{\ell}(i)\}_{i \in [n]}\right\}_{\ell \in [t] \setminus \{y,z\}}\right)
&\leq 2 \exp\left(-\frac{(1-\delta) n\tp \theta^2 (1-o(1))}{2 \hsig_k^{2s}}\right).
\end{align*}

We will choose $\theta = \hsig_k^s ((1-\delta) n \tp)^{-1/2} \log^{1/2}(trn^2)$, such that with probability $1 - \frac{2(1+o(1))}{tr n^2}$, 
\[|Y_{a,j} - e_k^T \hSig^{s+1} \tQ_y^T e_a| \leq \frac{\hsig_k^s \log^{1/2}(trn^2)}{((1-\delta)n\tp)^{1/2}},\]
which implies event $\cA^2_{a,j}(\delta)$ holds.
\end{new-proof}

\subsection{Concentration of Inner Product Statistic} \label{sec:innerprod}
\begin{new-proof}{Lemma \ref{lem:inner_prod}}
Recall that we assume $T_2^{obs}$ is a fresh data sample (alternatively this assumption can be removed by sample splitting instead).
Let us define the $\text{shrink}_{\phi}$ operator to be
\[\text{shrink}_{\phi}(x) = \begin{cases} \phi &\text{ if } x > \phi \\ x &\text{ if } x \in [-\phi,\phi] \\ -\phi &\text{ if } x < -\phi\end{cases}.\]

Without loss of generality assume $s$ is even (the same argument follows for odd $s$). Let us define 
\begin{align*}
X_{ij} &= N_{a,s}(i) N_{a',s+1}(j) \left(\sum_{\bh \in \cI_{yz}(i,j)} T^{obs}_2(\bh) \prod_{\ell \in [t] \setminus \{y,z\}} W_{\ell}(h_{\ell})\right) \\
\tX_{ij} &= N_{a,s}(i) N_{a',s+1}(j) ~\text{shrink}_{\phi}\left(\sum_{\bh\in \cI_{yz}(i,j)} T^{obs}_2(i,j,h) \prod_{\ell \in [t] \setminus \{y,z\}} W_{\ell}(h_{\ell})\right).
\end{align*}

The statistic $D(a,a')$ as defined in \eqref{eq:dist_D} can be constructed as sums of $X_{ij}$. We will show that with high probability $X_{ij} = \tX_{ij}$, and additionally $\left|\E{X_{ij}} - \E{\tX_{ij}}\right|$ is small. As $|T^{obs}(\bh)| \leq 1$ and $|W_{\ell}(h_{\ell})| \leq 1$ by our model assumptions, $|\sum_{\bh \in \cI_{yz}(i,j)} T^{obs}_2(\bh)  \prod_{\ell \in [t] \setminus \{y,z\}} W_{\ell}(h_{\ell})| \leq |\Omega_2 \cap \cI_{yz}(i,j)|$. It follows that $X_{ij} = \tX_{ij}$ whenever $|\Omega_2 \cap \cI_{yz}(i,j)| \leq \phi$. The difference between their expected values can be expressed as
\begin{align*}
	\left|\E{X_{ij}} - \E{\tX_{ij}}\right|
	&\leq \Ind{i \in \cS_{a,s}, j \in \cS_{a',s+1}}\E{\Ind{|\Omega_2 \cap \cI_{yz}(i,j)| > \phi} (|\Omega_2 \cap \cI_{yz}(i,j)| - \phi)},
\end{align*}
where $|\Omega_2 \cap \cI_{yz}(i,j)|$ is distributed as Bernoulli$(n^{t-2},p)$. 

For the setting where $\kappa \geq 1$, by Chernoff's bound,
\begin{align*}
\Prob( |\Omega_2 \cap \cI_{yz}(i,j)| > \phi) \leq \exp\left(-\frac{\phi - pn^{t-2}}{3}\right)
\end{align*}
and 
\begin{align*}
\E{\Ind{|\Omega_2 \cap \cI_{yz}(i,j)| > \phi} (|\Omega_2 \cap \cI_{yz}(i,j)| - \phi)} 
&= \sum_{g = 1}^{\infty} \Prob( |\Omega_2 \cap \cI_{yz}(i,j)| \geq \phi + g) \\
&\leq \sum_{g = 1}^{\infty} \exp\left(-\frac{\phi + g - pn^{t-2}}{3}\right) \\\
&\leq 3 \exp\left(-\frac{\phi - pn^{t-2}}{3}\right).
\end{align*}
If $\kappa=1$, we choose $\phi = 24\log(n)$ such that $\Prob( |\Omega_2 \cap \cI_{yz}(i,j)| > \phi)$ and $	\left|\E{X_{ij}} - \E{\tX_{ij}}\right|$ are bounded by $O(n^{-8})$. If $\kappa > 1$, then we choose $\phi = 2 pn^{t-2}$ such that the $\Prob( |\Omega_2 \cap \cI_{yz}(i,j)| > \phi)$ and $\left|\E{X_{ij}} - \E{\tX_{ij}}\right|$ decays exponentially in $n$, i.e. bounded by $\exp(-n^{\kappa-1})$.

For the ultra-sparse setting where $\kappa \in (0,1)$, we use a different argument as Chernoff's bound is not strong enough.
\begin{align*}
	\Prob(|\Omega_2 \cap \cI_{yz}(i,j)| > \phi) &= \sum_{g=\phi+1}^{n^{t-2}} \binom{n^{t-2}}{g} p^g (1-p)^{n^{t-2}-g} \\
	&< (1-p)^{n^{t-2}} \sum_{g=\phi+1}^{n^{t-2}} \left(\frac{p n^{t-2}}{1-p}\right)^g \\
	&< (1-p)^{n^{t-2}} \left(\frac{p n^{t-2}}{1-p}\right)^{\phi + 1} \left(1 - \frac{p n^{t-2}}{1-p}\right)^{-1} \\
	&< \frac{(1-p)^{n^{t-2}+1}}{1 - p (n^{t-2} + 1)} \left(\frac{p n^{t-2}}{1-p}\right)^{\phi + 1} \\
	&= (p n^{t-2})^{\phi + 1} (1 + o(1)) = n^{-(1-\kappa)(\phi + 1)} (1 + o(1)).
\end{align*}
Similarly, 
\begin{align*}
	\E{\Ind{|\Omega_2 \cap \cI_{yz}(i,j)| > \phi} (|\Omega_2 \cap \cI_{yz}(i,j)| - \phi)}
	&= \sum_{g=\phi+1}^{n^{t-2}} (g - \phi) \binom{n^{t-2}}{g} p^g (1-p)^{n^{t-2}-g} \\
	&\leq (1-p)^{n^{t-2}} \sum_{g=1}^{n^{t-2} - \phi} g \left(\frac{p n^{t-2}}{1-p}\right)^{g+\phi} \\
	&\leq (1-p)^{n^{t-2}} \left(\frac{p n^{t-2}}{1-p}\right)^{\phi+1} \left(1-\frac{p n^{t-2}}{1-p}\right)^{-2}\\
	&= \frac{(1-p)^{n^{t-2}+2}}{(1-p (n^{t-2}+1))^{2}} \left(\frac{p n^{t-2}}{1-p}\right)^{\phi+1}.
\end{align*}
We choose $\phi = \lceil 8 \ln(n)/\ln((p n^{t-2})^{-1}) \rceil = \lceil 8 /(1-\kappa)\rceil$, so that  $\Prob( |\Omega_2 \cap \cI_{yz}(i,j)| > \phi)$ and $\left|\E{X_{ij}} - \E{\tX_{ij}}\right|$ are bounded by $O(n^{-8})$.
Therefore, by union bound, $\Prob(\cup_{ij}\{X_{ij} \neq \tX_{ij}\}) \leq \sum_{ij} \Prob( |\Omega_2 \cap \cI_{yz}(i,j)| \geq \phi) \leq n^{-6}$. 

Next, we show that $\tilde{X}_{ij}$ concentrates around $\E{\tilde{X}_{ij}}$. As $\tX_{ij}$ results from shrinking values of $X_{ij}$ towards zero, the variance of $\tX_{ij}$ is bounded by the variance of $X_{ij}$. Conditioned on $\cF_{a,s}$, $\cF_{a',s+1}$ and $\left\{\{x_{\ell}(i)\}_{i \in [n]}\right\}_{\ell \in [t] \setminus \{y,z\}}$, i.e. the latent variables for modes $t \setminus \{y,z\}$,
\begin{align*}
	&\Var[X_{ij} ~|~ \cF_{a,s}, \cF_{a',s+1}, \left\{\{x_{\ell}(i)\}_{i \in [n]}\right\}_{\ell \in [t] \setminus \{y,z\}}] \\
	&= N_{a,s}^2(i) N_{a',s+1}^2(j) \sum_{\bh\in \cI_{yz}(i,j)} \sum_{\bh'\in \cI_{yz}(i,j)} \Cov[T^{obs}_2(\bh), T^{obs}_2(\bh')] \\
	&= N_{a,s}^2(i) N_{a',s+1}^2(j) \sum_{\bh\in \cI_{yz}(i,j)} \Var[T_2^{obs}(\bh)]  \\
	&\leq p n^{t-2} N_{a,s}^2(i) N_{a',s+1}^2(j).
\end{align*}
By independence of $\tX_{ij}$ and given that $\|N_{a,s}\|_{\infty} \leq 1$ and $\|N_{a',s+1}\|_{\infty} \leq 1$,
\[\Var\left[\sum_{i,j} \tX_{ij}\right] \leq p n^{t-2} |\cS_{a,s}| |\cS_{a',s+1}|.\]
By construction of the shrink operator, $|\tX_{ij}| \leq \phi$. 
As a result, it follows by Bernstein's inequality that 
\begin{align*}
&\Prob\left(\left. \Big|\sum_{i,j} (\tX_{ij} - \E{\tX_{ij}})\Big| \geq \theta ~\right|~ \cF_{a,s}, \cF_{a',s+1}, \left\{\{x_{\ell}(i)\}_{i \in [n]}\right\}_{\ell \in [t] \setminus \{y,z\}}\right) \\
&\qquad\leq 2 \exp\left(-\frac{3\theta^2}{6 p n^{t-2} |\cS_{a,s}| |\cS_{a',s+1}| + 2 \phi \theta}\right)
\end{align*}
We choose $\theta = \max\left(2 \log^{1/2}(t n^3) (p n^{t-2} |\cS_{a,s}| |\cS_{a',s+1}|)^{1/2}, \frac{4\phi \log(tn^3)}{3}\right)$, such that the above probability is bounded above by $2/tn^3$. Conditioned on $\cA^1_{a,s}(\delta,C), \cA^1_{a',s}(\delta,C)$, it follows that $|\cS_{a,s}| |\cS_{a',s+1}| = \Theta(n)$ such that $p n^{t-2} |\cS_{a,s}| |\cS_{a',s+1}| = \Theta(n^{1+\kappa})$.

Conditioned on $\cF_{a,s}$, $\cF_{a',s+1}$ and $\left\{\{x_{\ell}(i)\}_{i \in [n]}\right\}_{\ell \in [t] \setminus \{y,z\}}$,
\begin{align*}
&\E{X_{ij} ~\left|~ \cF_{a,s}, \cF_{a',s+1}, \left\{\{x_{\ell}(i)\}_{i \in [n]}\right\}_{\ell \in [t] \setminus \{y,z\}}\right.}  \\
&= N_{a,s}(i) N_{a',s+1}(j) \left(\sum_{\bh \in \cI_{yz}(i,j)} \E{T^{obs}_2(\bh)}  \prod_{\ell \in [t] \setminus \{y,z\}} W_{\ell}(h_{\ell}) \right) \\
&= N_{a,s}(i) N_{a',s+1}(j) p \sum_{\bk} \Lambda(\bk) q_{yk_y}(x_y(i)) q_{zk_z}(x_z(j)) \prod_{\ell \in [t] \setminus \{y,z\}} \sum_{h \in [n]} q_{\ell k_{\ell}}(x_{\ell}(h)) W_{\ell}(h) \\
&= p N_{a,s}(i) N_{a',s+1}(j) \sum_{a,b \in [r]^2} q_{ya}(x_y(i)) q_{zb}(x_z(j)) \hat{\Lambda}(a,b) n^{t-2} \\
&= p n^{t-2} N_{a,s}(i) N_{a',s+1}(j) e_i^T Q_y \hat{\Lambda} Q_z^T e_j \\
&= p n^{t-2} N_{a,s}(i) N_{a',s+1}(j) e_i^T \tQ_y \hat{\Sigma} \tQ_z^T e_j
\end{align*}
such that 
\begin{align*}
\frac{1}{p n^{t-2} |\cS_{a,s}| |\cS_{a',s+1}|} \sum_{i,j} \E{X_{ij} ~\left|~ \cF_{a,s}, \cF_{a',s+1}, \left\{\{x_{\ell}(i)\}_{i \in [n]}\right\}_{\ell \in [t] \setminus \{y,z\}} \right.}
&= \tN_{a,s}^T \tQ_y \hSig \tQ_z^T \tN_{a',s+1}.
\end{align*}

For sufficiently large $n$, conditioned on $\cA^1_{a,s}(\delta,C), \cA^1_{a',s}(\delta,C), \cA^2_{a,s}(\delta), \cA^2_{a',s+1}(\delta)$ for constants $\delta$ and $C$, it holds that with probability $1 - \frac{2}{t n^3}-\frac{1}{n^6}$,
\begin{align*}
&\left|D(a,a') - \tN_{a,s}^T \tQ_y \hSig \tQ_z^T \tN_{a',s+1}\right|  \\
&\leq  \max\left(\frac{2 \log^{1/2}(t n^3)}{(p n^{t-2} |\cS_{a,s}| |\cS_{a',s+1}|)^{1/2}}, \frac{4\phi \log(tn^3)}{3 p n^{t-2} |\cS_{a,s}| |\cS_{a',s+1}|}\right) (1+o(1)) \\
&= \Theta\left(\max\left(\frac{\log^{1/2}(t n^3)}{n^{(1+\kappa)/2}}, \frac{\log(tn^3)}{n^{2}}\right)\right),
\end{align*}
where the last equality comes from plugging in the choice of $\phi$ as 
\begin{align*}
	\phi = \begin{cases}
		\lceil 8 \ln(n)/\ln((p n^{t-2})^{-1})\rceil &\text{ if } pn^{t-2} = n^{\kappa-1} \text{ for }\kappa \in (0,1) \\
		24 \log(n) &\text{ if } pn^{t-2} = \Theta(1) \\
		2pn^{t-2} &\text{ if } pn^{t-2} = n^{\kappa-1} \text{ for }\kappa > 1,
	\end{cases}
\end{align*}
and verifying the above holds for each case of $\kappa < 1$, $\kappa = 1$, and $\kappa > 1$.


\end{new-proof}

\end{document}